\documentclass[10pt,twocolumn,letterpaper]{article}

\usepackage[pagenumbers]{cvpr} % To force page numbers, e.g. for an arXiv version

\makeatletter
\@namedef{ver@everyshi.sty}{}
\makeatother
\usepackage{graphicx}
\usepackage[export]{adjustbox}
\usepackage{amsmath}
\usepackage{amssymb}
\usepackage{booktabs}
\usepackage{tabularx}
\usepackage{array}
\usepackage{float}
\usepackage{wrapfig}
\usepackage{multirow}
\usepackage{bm}
\usepackage{tikz,pgfplots}
\usetikzlibrary{spy}

\usepackage[pagebackref,breaklinks,colorlinks]{hyperref}
\newcommand{\vt}[1]{\mathbf{#1}}
\newcommand{\mat}[1]{\bm{#1}}%\mathbf{#1}} % matrices (usually uppercase)

\newcommand{\I}{\mathcal{I}}

\newcommand{\SDF}{\textbf{MLP}_{\text{SDF}}}
\newcommand{\COLOR}{\textbf{MLP}_{\text{COLOR}}}

\usepackage{xcolor}

\usepackage{ulem}
\usepackage{afterpage}

\newcommand*{\affaddr}[1]{#1} % No op here. Customize it for different styles.
\newcommand*{\affmark}[1][*]{\textsuperscript{#1}}
\usepackage[capitalize]{cleveref}
\crefname{section}{Sec.}{Secs.}
\Crefname{section}{Section}{Sections}
\Crefname{table}{Table}{Tables}
\crefname{table}{Tab.}{Tabs.}

\def\cvprPaperID{2781} % *** Enter the CVPR Paper ID here
\def\confName{CVPR}
\def\confYear{2024}

\begin{document}

\title{Coloring the Past: Neural Historical Buildings Reconstruction \\ from Archival Photography}
% david \orcidID{0000-0002-5639-2547}
% ferdinand \orcidID{0000-0002-2456-9731}
% \author{Dávid Komorowicz \quad Lu Sang \quad Ferdinand Maiwald \quad Daniel Cremers\\
% Technical University of Munich \quad Institute of Photogrammetry and Remote Sensing, Technical University Dresden, Germany \\ 
% Computer Vision Group\\
% {\tt\small david.komorowicz, lu.sang, cremers@tum.de \quad ferdinand.maiwald@tu-dresden.de}

% Chair for Digital Humanities (Images/Objects), Friedrich Schiller University Jena, Germany


\author{
Dávid Komorowicz$^{*}$\affmark[1,3] \quad Lu Sang$^{*}$\affmark[1] \quad Ferdinand Maiwald\affmark[2] \quad Daniel Cremers\affmark[1] \vspace*{0.3cm} \\
\affaddr{\affmark[1]Technical University of Munich\\Computer Vision Group}\\
\begin{tabular}[h]{cc}
    \affaddr{\affmark[2]Dresden University of Technology} & \affaddr{\affmark[3]Friedrich Schiller University Jena} \\
    Institute of Photogrammetry and Remote Sensing & Chair for Digital Humanities
\end{tabular}\\
% \affaddr{\affmark[2]Technical University Dresden\\Institute of Photogrammetry and Remote Sensing} \quad
% \affaddr{\affmark[3]Friedrich Schiller University Jena\\Chair for Digital Humanities} \\
{\tt\small david.komorowicz, lu.sang, cremers@tum.de \quad ferdinand.maiwald@tu-dresden.de}
}

% \maketitle

% \begin{figure*}[!th]
%   \centering
%   \includegraphics[width=0.8\textwidth]{imgs/teaser_placeholder.png}
%   \caption{Your teaser image caption goes here.}
% \end{figure*}

% \todo{TEASER, add your professor's name (no worries, we only need this in the future)}
% \begin{itemize}
%     \item \todo{title color or colour?}
%     \item \todo{teaser}
%     \item \todo{check all the text I wrote, if they are correct.}
%     \item \todo{add citations.}
%     \item \todo{Check the abbreviations. Every abbreviation should only be mentioned once and then be consequently used (e.g. SfM)}
%     \item \todo{grayscale, etc should be written the same way everywhere}
%     \item \todo{cite geo neus for sdf loss, we show in the ablation that dense pcd is better than sparse pcd}
%     \item \todo{agisoft cite old version}
%     % \item \todo{rewrite related work}
% \end{itemize}

% teaser> -historical images -> colmap pcd with few cameras -> colored mesh -> close up mesh/normal
% % -2 column or 1 column
% % -theater
\twocolumn[{%
\renewcommand\twocolumn[1][]{#1}%
\maketitle
\begin{center}
\vspace{-0.5em}
% \begin{figure}
\renewcommand\fbox{\fcolorbox{orange}{white}}
\setlength{\tabcolsep}{2pt}
\newcolumntype{Y}{>{\centering\arraybackslash}p{0.28\linewidth}}
\begin{tabular}{YcYY}
     \includegraphics[width=\linewidth]{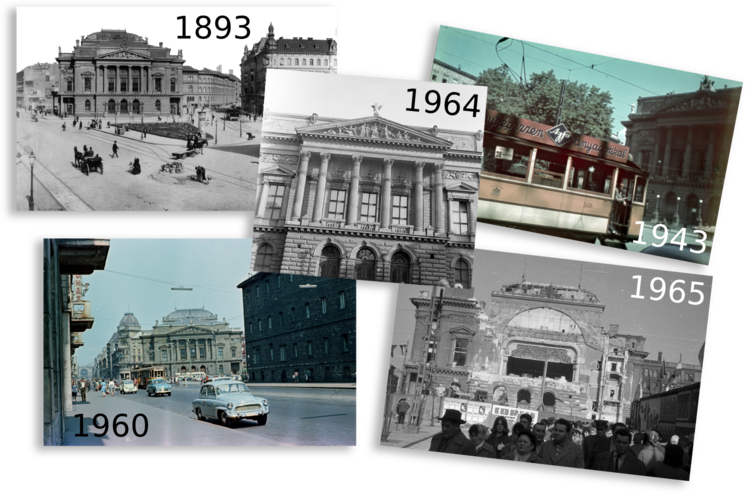} &
     \centering
     \includegraphics[width=0.05\linewidth, margin=0pt 80pt 0pt 0pt, valign=m]{imgs/arrow.png}
     &
     \includegraphics[width=\linewidth]{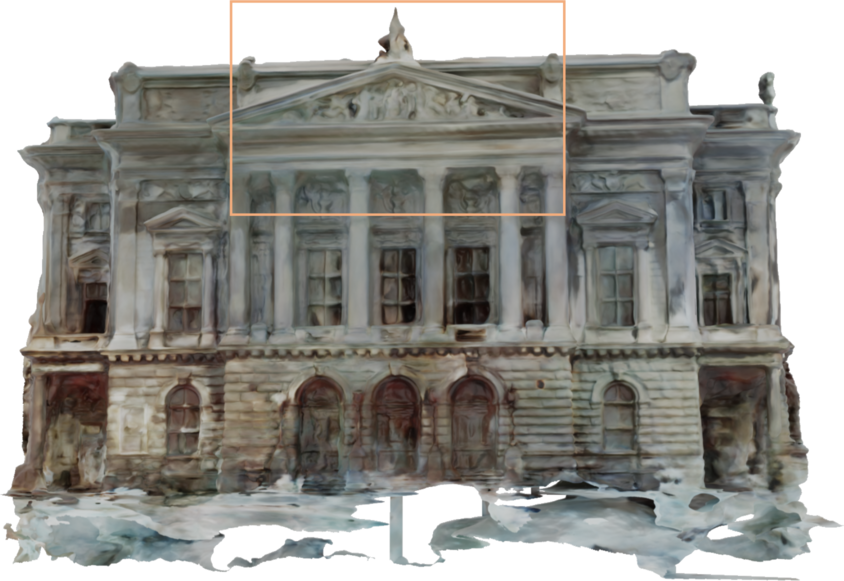}& 
    {%
    \setlength{\fboxsep}{0pt}%
    \setlength{\fboxrule}{1pt}%
    \fbox{\includegraphics[width=0.9\linewidth]{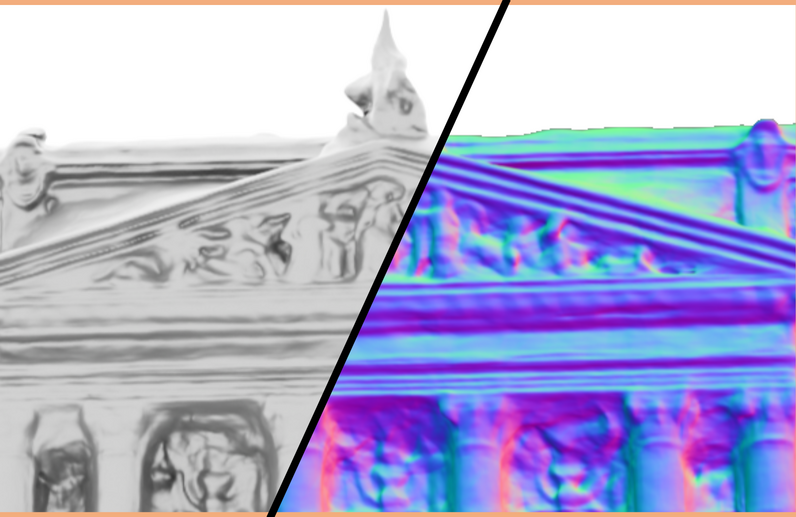}}%
    }%
\end{tabular}
\vspace{-1cm}
   \captionof{figure}{\textbf{Coloring the Past}: Reconstructing high-quality historical buildings from limited data and recovering color appearance from a majority of gray-scale images. The figure shows photographs of the Hungarian National Theater over a long time period (left), reconstructed mesh with color (middle), and shaded mesh + normal map (right).}
    \label{fig:teaser}
% \end{figure}
\end{center}%
}]

\let\thefootnote\relax\footnote{$^{*}$ These authors contributed equally.}
%%%%%%%%% ABSTRACT
\begin{abstract}
Historical buildings are a treasure and milestone of human cultural heritage. Reconstructing the 3D models of these building hold significant value. The rapid development of neural rendering methods makes it possible to recover the 3D shape only based on archival photographs. However, this task presents considerable challenges due to the limitations of such datasets. Historical photographs are often limited in number and the scenes in these photos might have altered over time. The radiometric quality of these images is also often sub-optimal. To address these challenges, we introduce an approach to reconstruct the geometry of historical buildings, employing volumetric rendering techniques. We leverage dense point clouds as a geometric prior and introduce a color appearance embedding loss to recover the color of the building given limited available color images. We aim for our work to spark increased interest and focus on preserving historical buildings. Thus, we also introduce a new historical dataset of the Hungarian National Theater, providing a new benchmark for the reconstruction method.

% Multi-view reconstruction holds significant potential in many application, such as reconstructing scenes and buildings that may no longer exist. \ferdinand{first sentence needs to be improved} However, this task presents considerable challenges due to the limitations on such dataset. Firstly, historical photographs are often hard to collect and the scenes in these photos might have altered over time which introduce inconsistencies across images. %\david{pcd helps preserves these details}
% Secondly, the quality and quantity of these images are usually sub-optimal. We introduce an approach for reconstructing the geometry of historical buildings using archival images, employing volumetric rendering techniques. We leverage dense point clouds as a geometric prior which helps to preserve geometry details and introduce a color appearance embedding loss. This loss is designed to maximize the utility of the limited color images in datasets which are mainly composed of gray-scale images, thereby enabling the recovery of colored meshes. We aim for our work to spark increased interest and focus on the reconstruction of historical buildings. To further follow this goal, we introduce a new historical dataset of the Hungarian National Theater, providing a valuable resource for evaluating reconstruction methods.
\end{abstract}

\section{Introduction}\label{sec:intro}
% \lu{makes it more readable}
% \ferdinand{I made several changes in the introduction+related works and added citations.} \lu{cool, thanks!}
Historical buildings embody the unique characteristics of cultural heritage and are seen as landmarks that connect people across time and countries. 
They make history tangible, yet they are vulnerable to temporal and man-made changes.
It is one of the main goals of UNESCO to protect and preserve our cultural heritage which becomes possible because of the rapid development of 3D technologies \cite{SkublewskaPaszkowska2022}.
%Historical buildings stand as \textbf{tangible narratives} of cultural heritage and artistic endeavor, yet they are persistently vulnerable to the \textbf{ravages} of time and neglect.
% tips from Niessner :) https://twitter.com/MattNiessner/status/1724795600456310844
%Digitization of the 3D shape and views of historical buildings have a significant meaning in many aspects. 

(Historical) images are usually processed in a Structure-from-Motion (SfM) workflow to estimate the intrinsic and extrinsic camera parameters and a dense 3D scene is conventionally reconstructed using multi-view stereo (MVS) \cite{colmap}.
With the advent of representing 3D scenes as Neural Radiance Fields (NeRF)~\cite{mildenhall2021nerf}, the published framework facilitates the synthesis of photo-realistic images from novel viewpoints, using a volumetric scene representation learned from sparse and unstructured 2D images. 
The ability of NeRF to interpolate and extrapolate from image data introduces an unprecedented potential for reconstructing historical buildings using mostly historical photographs as a source \cite{jena}.\par %that might only persist in the \textbf{annals} of photographic history. \par

However, compared to the reconstruction of modern buildings or scenes, recovering synthetic views and the 3D shape of historical buildings comes with several limitations. 
A significant issue lies in the scarcity and the often-compromised quality of available input data \cite{Maiwald2022a,Morelli2022,Knuth2023}. 
Many historical sites are documented solely through antiquated photographs, captured with obsolete equipment that takes images with difficult radiometric properties such as blurriness, lack of color, or absence of accurate camera parameters \cite{Maiwald2019,Poli2020,Chumachenko2020finnishworldwar}. 
Additionally, only a limited number of photos can be found and used, coupled with the extensive temporal spread across which they were taken, which means that inconsistencies are in both the structural state of the buildings and in the photographic records themselves \cite{Maiwald2019,Farella2022}. 

In this paper, we propose a method that tackles 3D reconstruction and colored view syntheses for historical buildings leveraging the sparse and low-quality input images.~\cref{fig:teaser} shows the reconstruction results of our method. Given a historical image collection dominated by gray-scale images over different time, we are able to recover the colored 3D mesh of the building. 
We also would like to raise interest in the topic of historic monument reconstruction and the use of historical photographs in the 3D reconstruction community. \par
In summary, our contributions are as follows:
\vspace{-0.3cm}
\begin{itemize}
    % \item view synthesis + 3D mesh
    \item We propose a method that is able to reconstruct satisfactory 3D geometry of historical buildings by leveraging sparse and low-quality images.
    \vspace{-0.3cm}
    \item We propose a color appearance embedding loss to obtain a color synthetic view when the majority of photos are gray-scale.
    \vspace{-0.3cm}
    \item Our method achieves better reconstruction results by incorporating already existing data.
    \vspace{-0.3cm}
    \item We publish a historical dataset that showcases a wide range of properties typically present in historical datasets.
    % \item experiment shows our results are sota?
\end{itemize}

% Cultural heritage reconstrution is a very important field, etc...

% , lost monuments limited number of imgaes

% Our contributions:

% \begin{itemize}
%     \item We publish a historical dataset which showcases a wide range of properties typically present in historical datasets.
%     \item We show how to incorporate already existing data to improve reconstuction results.
%     \item We show how to obtain a color appearance from a dataset where the majority of photos is grayscale.
%     \item Finally, we want to raise interest in the topic of historic monument reconstruction in the 3d reconstitution community. \lu{I added in text}
% \end{itemize}
%------------------------------------------------------------------------
\section{Related Work}\label{sec:related_works}
\paragraph{Multi-view 3D reconstruction} %with stereo fusion, accurate geometry at points, holes at textureless regions noisy surface because separate steps
Reconstructing the underlying 3D geometry from multiple images from different viewpoints is a long-studied problem in the computer vision field. 
Conventional multi-view stereo (MVS) methods~\cite{agrawal2001,bleyer2011patchmatch,Bonet1999PoxelsPV,pixelwise,leroy2018shape, sang2023high,Sommer2022}, consider matching geometry priors such as depth~\cite{pixelwise} or using voxel as surface representation and project points back to the image to refine the geometry~\cite{sang2023high, Sommer2022}. 
A limitation of traditional approaches is, that they often rely on discrete representations like depth maps, or voxel volumes to model surface space. 
This approach can lead to substantial memory usage when dealing with large scenes. 
Additionally, the process of finding correspondences is generally vulnerable to noise, which can affect the accuracy of the reconstruction. 
\textbf{Learning-based} multi-view methods~\cite{neus2,martinbrualla2020nerfw,yariv2021volume}, usually use networks to learn or extract features from color images with~\cite{dsnerf, Yu2022MonoSDF,fu2022geoneus} or without~\cite{yariv2021volume, neus2} geometry priors. 
They are more robust to noise and generalize across different scenes and objects. 
However, they require large amounts of training data, which can be computationally intensive and may not be available for historical datasets.
% Nerfacto~\cite{nerfstudio} uses 2 hash-encoded fused MLPs as proposal networks to learn surface intersection points accelerate sampling. \david{please check}
% few images, no explicit geometric representation, not feasible to render in real time.
% which are suitable for synthesing novel-view images but cannot  % reason
% produce high-quality surface reconstrucions. In contrast, our approach models the scene with a surface-based representation and directly produces smooth and accurate 3D meshes.
\vspace{-0.4cm}
\paragraph{Surface representation scheme} During 3D reconstruction, the choice of surface representation method is crucial. Traditional approaches often utilize discrete methods like point clouds, polygon meshes, or discrete Signed Distance Fields (SDF)~\cite{Sommer2022, sang2023high, newcombe2011}.
On the other hand, learning-based techniques, aided by neural networks, can employ continuous surface representation methods. The most commonly used property is density~\cite{mildenhall2021nerf, martinbrualla2020nerfw}, which indicates the transparency value of the given point.
However, this approach, while effective for view synthesis, often fails in accurately reconstructing geometry. It calculates an integrated opacity value along a ray, rather than modeling an explicit surface point. Alternatively, continuous SDF, another form of surface representation, offers a more precise approach. It calculates the distance from any point in space to the nearest surface, indicating whether the point is inside (negative) or outside (positive) the object. This representation enables SDF-based methods~\cite{yariv2021volume, neus2, sun2022neuconw}, more accurate identification of surfaces, and better handling of occlusions and view-dependent effects compared to volume-only-based methods. NeuS~\cite{wang2021neus}, for instance, uses a differentiable rendering framework that combines SDF with a radiance field, resulting in high-quality images that capture fine geometric details. Neus-Facto~\cite{Yu2022SDFStudio} propose a network without surface-guided sampling and geometry prior.
% MonoSDF~\cite{Yu2022MonoSDF} takes a unique approach by using a single image to predict a depth map and a normal map. This is then employed as a geometric prior for refining initial geometry estimates. The authors demonstrate that incorporating normal and depth information provides complementary insights into the scene. However, the method needs the rays hitting the surface which is not always the case for outdoor scenes, which leads to the method being sensitive to initialization. Moreover, MonoSDF~\cite{Yu2022MonoSDF} only works in indoors because it needs surface intersection and linear scale assumption (sky). 
GeoNeus~\cite{fu2022geoneus} uses a sparse point cloud to supervise the SDF and a Photometric consistency loss. This is not suitable for historical imagery due to large illumination and appearance inconsistencies between images.

\vspace{-0.4cm}
\paragraph{Historical building dataset} The use of historical public and press photography showing terrestrial scenes is still a rather rare scenario for complex 3D reconstruction \cite{Falkingham2014,Beltrami2019,Maiwald2022a}, whereas historical aerial images are already commonly and increasingly used in Structure-from-Motion workflows \cite{Feurer2018,Zhang2021,Farella2022,Knuth2023,Maiwald2023a}.
Historical terrestrial images are mainly used for special tasks such as horizon line detection \cite{MikolkaFloery2021}, photographer recognition\cite{Chumachenko2020finnishworldwar}, and building height estimation \cite{Farella2021}.
Other approaches focus on the integration of historical images into further existing data such as terrestrial laser scanning \cite{Bitelli2017,Bevilacqua2019}, and contemporary photographs \cite{Maiwald2017a,Kalinowski2021}.
When working with historical images, the standard 3D reconstruction workflow is usually interrupted after sparse point cloud creation and camera pose estimation because conventional MVS strategies fail to generate reasonable surface representations \cite{Maiwald2022a}. 
% In our study, we employ an SDF-based surface representation along with a differentiable rendering approach to address the challenges of 3D reconstruction and view synthesis, particularly focusing on historical buildings. \david{duplicate?}

To overcome the difficulties posed by gray-scale, low-quality, and sparse input images, we integrate a dense point cloud as a geometric prior and introduce a novel color appearance embedding loss. 
We are confident that our methodology holds substantial value in preserving the historical heritage of human culture. 
% \lu{check this part if it is correct.}

\section{Historical Dataset}\label{sec:dataset}
Reconstructing historical buildings based on archival photography provides significant value not only in the research area but also considering the protection and preservation of cultural heritage. However, historical images of the same building are often scattered in multiple archives with often unresolved copyrights and only a few historical datasets are available for research purposes. Thus, we introduce the \textbf{Hungarian National Theater dataset}. \par%of the Hungarian National Theater. 
% \lu{only one dataset include hungarian national theater? I thought you have dataset for several buildings?}
This dataset includes 229 images of the Hungarian National Theater directly released by us and another 136 images for which the access link is provided. Additionally, we provide a dense point cloud and camera poses which are registered using Structure from Motion (SfM). 
% 258 of which are registered in the Structure from Motion (SfM) including, 66 photographs during demolition. % We publish the images together with registered poses and the point cloud for users' convenience. 
All photos were taken between 1875-1965. During this period, the availability of color photography was limited. Thus, different from the modern building image datasets, the vast majority of photos are gray-scale (over $90\%$) and only a small portion is available in color (~\cref{tab:dataset}). Another significant difference is, that the building can slightly change over the span of decades which is why we provide the capture dates of the images. 

\begin{figure}[t!]
    \centering
    \includegraphics[width=.45\textwidth]{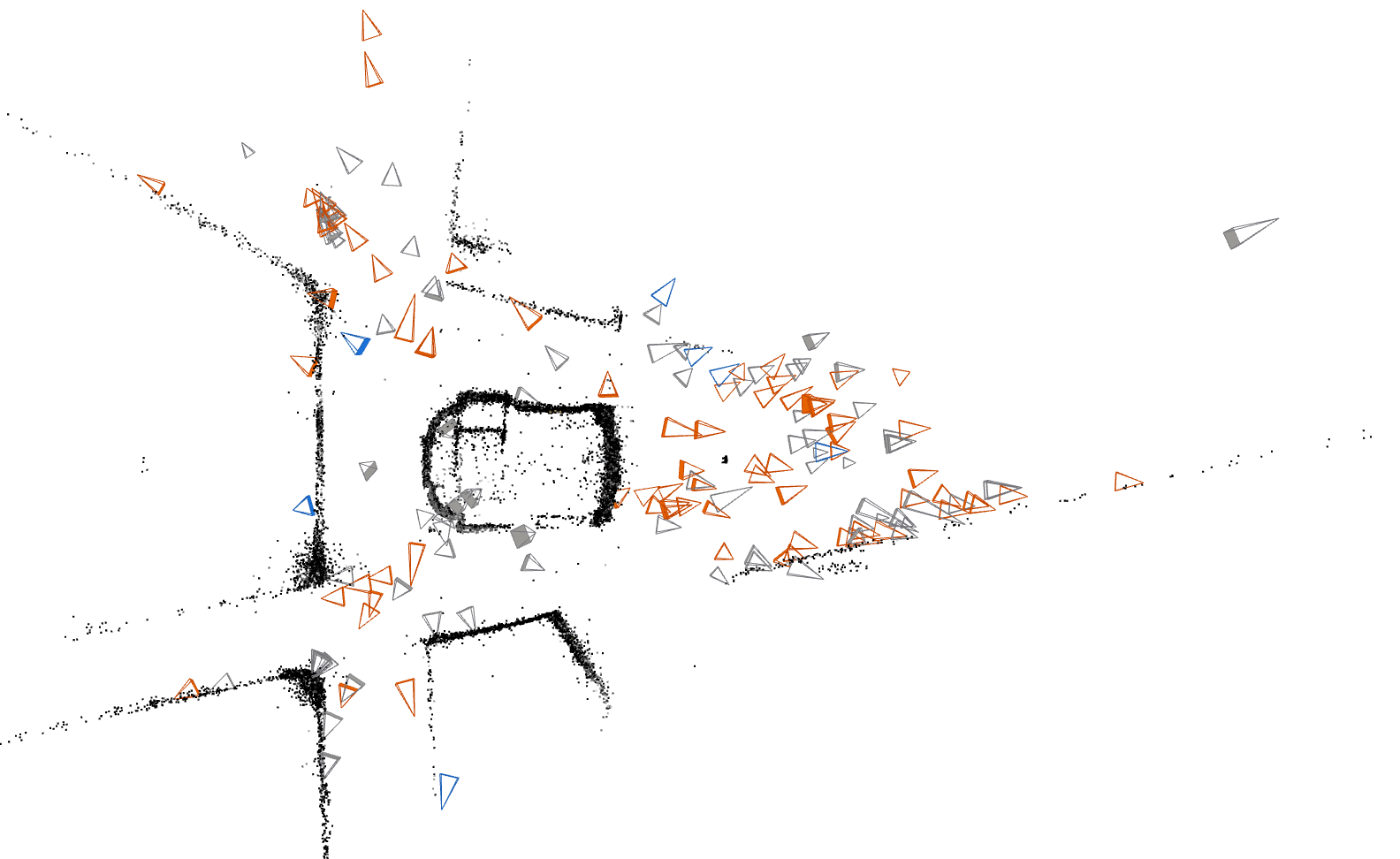}
    \caption[Top-down view of a reconstructed point cloud of the National Theater dataset. \textcolor{blue}{blue} cameras stand for validation views, \textcolor{orange}{orange} cameras are training images. \textcolor{lightgray}{gray} cameras are not free released images]{Top down of view of a reconstructed point cloud of the National Theater dataset: \textcolor{blue}{blue} cameras stand for validation views, \textcolor{orange}{orange} cameras are training images, \textcolor{lightgray}{gray} cameras are images that can be obtained via request\protect\footnotemark.}
    \label{fig:theater_topdown}
\end{figure}

\begin{figure}[t!]
\setlength{\tabcolsep}{2pt}
\newcolumntype{Y}{>{\centering\arraybackslash}p{0.32\linewidth}}
\begin{tabular}{YYY}
%   \rotatebox{90}{National Theater} &
% 82344
\includegraphics[width=\linewidth]{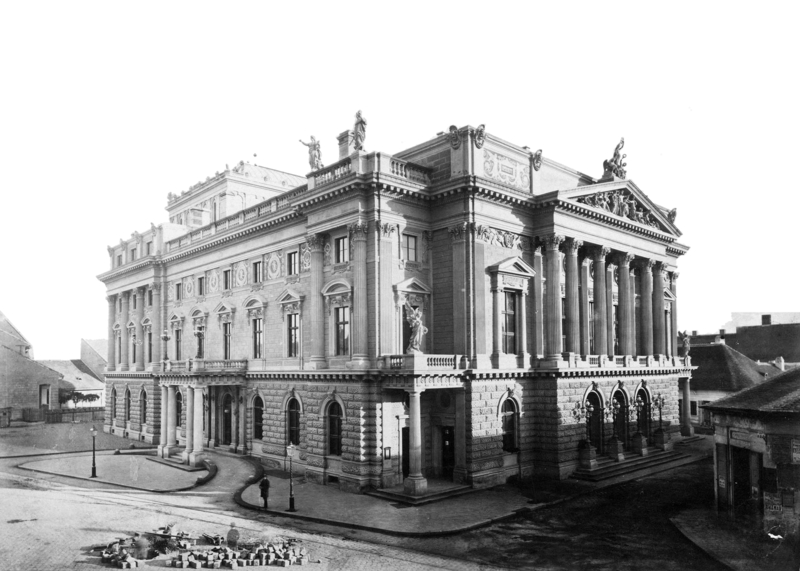}  & \includegraphics[width=\linewidth]{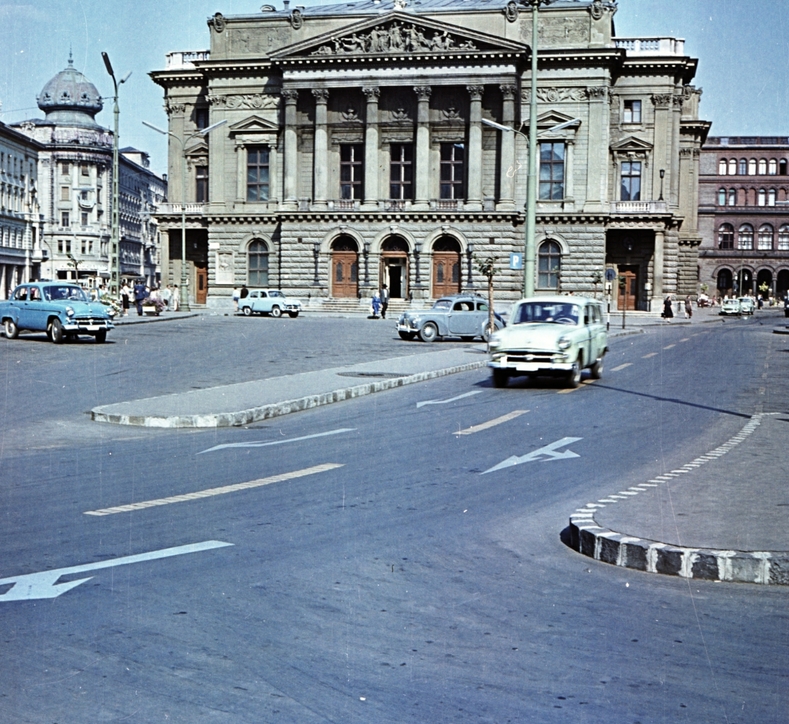}  & \includegraphics[width=\linewidth]{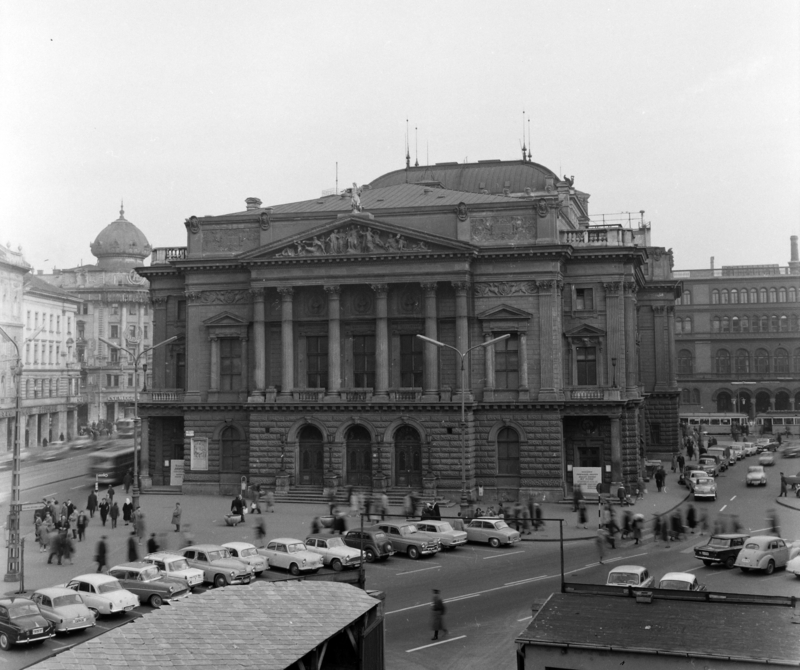}\\
% 1893 HU.BFL.XV.19.d.1.07.020
1875 & 1959 & 1964 \\
  
\end{tabular}
    \caption[fake caption]{Example images\protect\footnotemark from the \textbf{Hungarian National Theater} dataset.}
    \label{fig:dataset_historic_example}
\end{figure}

Besides its cultural significance to the Hungarian people, this dataset is a rare case of having a complete photo collection covering the whole area around an old building that is no longer present. All four sides appear in different numbers of images in the dataset. %It contains more images of the front view but less for the rear view. 
This makes the dataset suitable as a benchmark to evaluate the algorithms' performance regarding the number and quality of the input images. ~\cref{fig:theater_topdown} shows the reconstructed point cloud and estimated camera locations using SfM~\cite{colmap} from the National Theater dataset. \cref{fig:dataset_historic_example} shows example images of the facade across time.
%This enables us to benchmark algorithms on how well they work regarding the number of images by looking at each side (density).
\footnotetext{These images are private and can be accessed upon purchase. We will provide the link to these images.}

\begin{table}[h]
\centering
\begin{tabular}{lp{1cm}p{1cm}p{1cm}}
Dataset name & Total  & Color & Train \\
National Theater (Ours)     &   229   & 16 & 153  \\
Hotel International~\cite{jena} &     19   & 1   & 18 \\
Observatory~\cite{jena} &     37   & 3   & 33  \\
St. Michael Church~\cite{jena} &   17   & 0 & 16\\
% semperoper~\cite{dresden}                  & 131                \\
% stockholm                   & 37                 \\
% Brandenburg Gate (historic) &  1363 & 1/9 & 10/90                 
% Brandenburg Gate 10 &      10    &  10    & 1  \\
% Brandenburg Gate 50 &      50    &  10    & 5    \\  
% Brandenburg Gate 90 &      90    &  10    & 9  \\
\end{tabular}
\caption{Datasets statistics of National Theater dataset,  three historical datasets from~\cite{jena}.}
\label{tab:dataset}
\end{table}

% \begin{figure}[]
% \centering
%   \begin{tabular}{@{}cccc@{}}
%     theater & observatory & church & hotel \\
    
%     \includegraphics[width=.11\textwidth]{imgs/example_images/theater/fortepan_97192.jpg} &
%     \includegraphics[width=.11\textwidth]{imgs/example_images/observatory/2978.jpg} &
%     \includegraphics[width=.11\textwidth]{imgs/example_images/church/A6N_288.jpg} &
%     \includegraphics[width=.11\textwidth]{imgs/example_images/hotel/2456.jpg} 
%     \\
%     \includegraphics[width=.11\textwidth]{imgs/example_images/theater/fortepan_251236.jpg} &
%     \includegraphics[width=.11\textwidth]{imgs/example_images/observatory/2984.jpg} &
%     \includegraphics[width=.11\textwidth]{imgs/example_images/church/A6P_1291.jpg} &
%     \includegraphics[width=.11\textwidth]{imgs/example_images/hotel/2463.jpg} 
%     \\
%     \includegraphics[width=.11\textwidth]{imgs/example_images/theater/fortepan_102400.jpg} &
%     \includegraphics[width=.11\textwidth]{imgs/example_images/observatory/3010.jpg} &
%     \includegraphics[width=.11\textwidth]{imgs/example_images/church/A6P_339.jpg} &
%     \includegraphics[width=.11\textwidth]{imgs/example_images/hotel/2567.jpg} 
%     \\
%   \end{tabular}
%   \caption{Dataset example images}
%   \label{fig:dataset_historic_example}
% \end{figure}

\footnotetext{\scriptsize Fortepan by UVATERV/F\"{O}MTERV/Zsolt Pálinkás/Pál Breuer/Lajos Miklós and Budapest City Archives: HU.BFL.XV.19.d.1.05.103/HU.BFL.XV.19.d.1.07.020 \\ under CC-BY-SA-3.0}

% ~\cref{tab:dataset} shows the information of our dataset together with another historical dataset from~\cite{jena}. Compared to modern datasets, historical building datasets contain significantly fewer images, and most of them are not color images. These characteristics make historical building reconstruction challenging. Moreover, our new dataset contains more training and color images compared to datasets from~\cite{jena}, which enables different levels of algorithm performance evaluation.
% ~\cref{tab:dataset} presents data from our collection alongside a historical dataset from~\cite{jena}. When contrasted with contemporary datasets, those focusing on historical buildings typically have a limited number of images, predominantly in black and white. This limitation and lack of color contribute to the complexities of reconstructing historical buildings. Furthermore, our dataset offers more training images and more color images, compared to the collections mentioned in ~\cite{jena}. This enhancement facilitates a more diverse evaluation of a method's performance.

~\cref{tab:dataset} summarizes all information for the released historical dataset and three further historical datasets \cite{jena}. We use these four datasets to test our methods in \cref{sec:eva}. The first column shows the total number of images for the dataset, the second column provides the number of color images, and the third column shows the number of images that we actually use to train our model. Due to the quality of the images, not all of them are suitable. From ~\cref{tab:dataset} we can see that our dataset contains an order of magnitude more images in total and more color images as well.  
\section{Method}\label{sec:method}
The whole pipeline of our method is as follows. Given a set of images $\{\I_i\}$, for $i \in \{0,1,\dots, n\}$, we first resize the images to the same size, since the historical datasets typically contain images with varying resolution. Then, the corresponding extrinsic (poses) and intrinsic camera parameters are estimated using SfM~\cite{colmap}. We run a segmentation method, similar to~\cite{sun2022neuconw} to mask out irrelevant objects such as people and cars. We generate two kinds of point clouds, a \textbf{sparse} point cloud, directly using SfM~\cite{colmap} and a \textbf{dense} point cloud $\mathcal{P} = \{ \vt{x}_1, \vt{x}_2, \dots, \vt{x}_n\}$ using the estimated camera parameters by multiview stereo fusion~\cite{colmap}. We use the dense point cloud as geometry prior together with images to train an SDF-based differential rendering network with color appearance embedding loss to estimate the geometry.% and enable render views from unseen camera poses. 
\subsection{Backbones and Geometry Loss}
We build our method on top of NeusW~\cite{sun2022neuconw}. Our network architecture consists of two parts, an SDF net and a color prediction net. The SDF net estimates the signed distance value $d\in\mathbb{R}$ and a geometric feature $\vt{f}\in\mathbb{R}^{f_n}$, for $f_n$ is the dimension of the feature vector. Given point $\vt{x}\in\mathbb{R}^3$, the color prediction net outputs the rendered color $\vt{c}$. In detail, given points $\vt{x}$, viewing direction $\vt{v}\in\mathbb{S}^2$, we compute normal $\vt{n}=\nabla \SDF(\vt{x})$, 
%camera dependent appearance embedding vector $\vt{e_i}$ 
and a feature vector $\vt{f} \in \mathbb{R}^{f_n}$ with dimension $f_n$.
\begin{align}
    % \SDF: & \mathbb{R}^3 \to \mathbb{R} \\
    \label{eq:geo_net} (d,\vt{f}) &= \SDF(\vt{x}) \,, \\ 
    % \COLOR: &\mathbb{R}^3\times \mathbb{R}^3\times \mathbb{R}^{f_n} \to \mathbb{R}^3 \\
    \label{eq:color_net}\vt{c}_i &= \COLOR(\vt{x}, \vt{v}, \vt{n}, \vt{e}_i, \vt{f})\,.
\end{align}
where $\vt{e}_i$ are appearance embeddings corresponding to each input photo, optimized alongside the parameters of MLPs, see~\cite{sun2022neuconw} for more details.
We first initialize a voxel grid by the sparse point cloud similar to~\cite{sun2022neuconw}. For image $\I_i$ with camera center $\vt{o}$, we shoot a ray from its pixels. The ray $\vt{r}$ with direction $\vt{v}$ is $\{\vt{r}(s) = \vt{o} + \vt{v}s|s\geq0\}$. We pass the points along the ray to the SDF net to get the geometry feature $\vt{f}$ and then pass these points to the color net to get the color estimation. 
We reuse the SDF net for geometry loss as well. For image $\I_i$ where we sampled ray from, we find all points from the dense point cloud $\mathcal{P}$, which are visible from this image, denoted as $\mathcal{P}_i$. The geometry loss\cite{fu2022geoneus} is 
\begin{equation}
\label{eq:geometry_loss}
\mathit{l}_{g}(\vt{x}) = \lambda \frac{1}{|\mathcal{P}_i|}\sum_{\vt{x} \in \mathcal{P}_i}{|\SDF(\vt{x})|}\,,
\end{equation}
where $|\mathcal{P}_i|$ is the number of points in the point cloud and $\lambda$ is a learnable parameter. During training, we sample rays across multiple images for one batch and randomly choose one image to compute the geometry loss for the point cloud visible from that image.
The geometry loss ensures that the SDF net is guided by the \textbf{dense point cloud}. 

\begin{figure}[b!]
\centering
\begin{minipage}{0.45\linewidth}
        \centering
        \includegraphics[width=\linewidth]{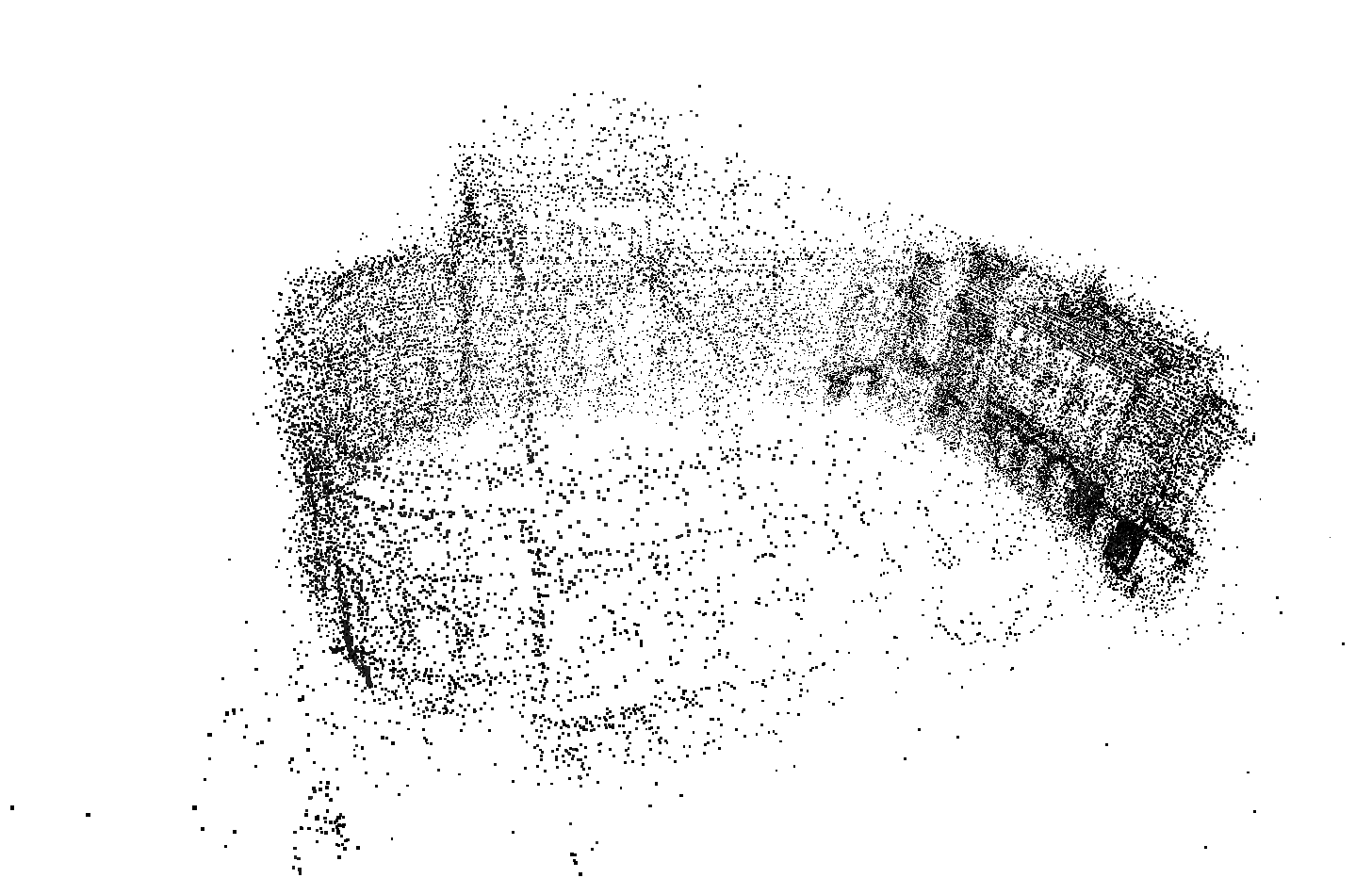} % First image
    \end{minipage}%
    \begin{minipage}{0.45\linewidth}
        \centering
        \includegraphics[width=\linewidth]{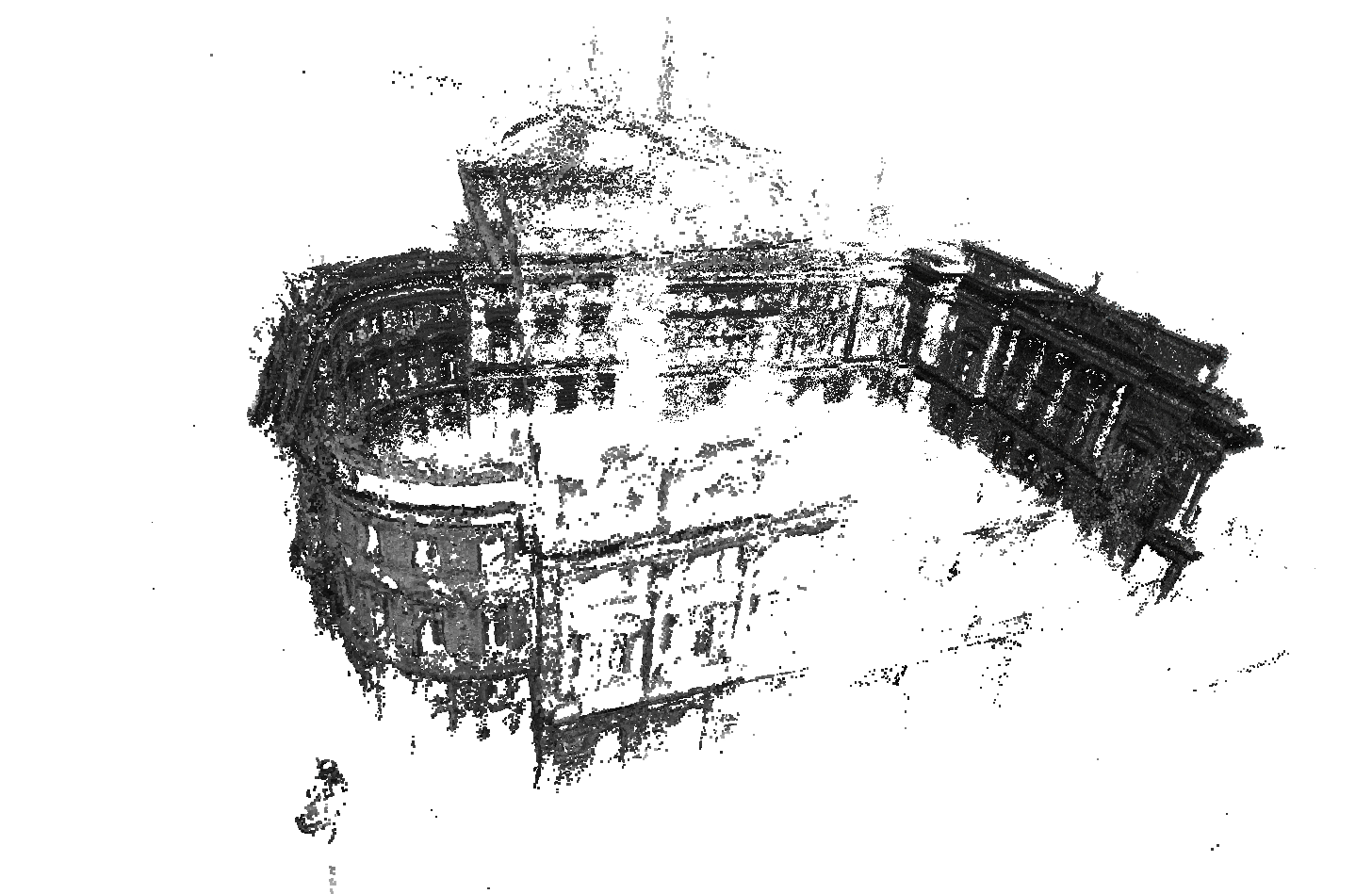} % Second image
    \end{minipage}
    % \centering
    % \includegraphics[width=0.45\textwidth]{imgs/ablation/gate_10/pcd.jpg_color.png}
    \caption{Comparison of sparse (left) and dense (right) point cloud generated by stereo fusion~\cite{colmap}.}
    \label{fig:sparse_dense_pcd}
\end{figure}

We use a dense point cloud instead of the sparse point cloud because we believe the dense point cloud provides complementary information, see~\cref{fig:sparse_dense_pcd}. Directly sampling at the dense point cloud points to optimize the SDF net allows us to bypass the ray marching procedure. In NeusW~\cite{sun2022neuconw} and our case, the sampling is directly dependent on the SDF values. Good geometry prior, \ie dense point cloud will benefit SDF estimation first, and the improved SDF will improve sampling again. %Since, after that, we follow the surface-guided sampling that uses SDF information similar to NeusW~\cite{sun2022neuconw}. \david{previous sentence is kind of duplicate} This is different from neus-facto~\cite{Yu2022SDFStudio}, which does not use geometry prior, and only depends on the color network.

\subsection{Color Appearance Embedding} \label{subsec:color}
To deal with the situation that most of the input images are available as gray-scale, and only a small portion provides color channels, we propose a color appearance embedding loss to recover color output. Previous methods treat gray-scale images as color by setting the three channels to equal values. This results in less-than-ideal appearance embedding and a gray-scale output. The rendered color for a ray $\vt{r}$ is 
\begin{equation}\label{eq:color_render}
    \mat{C}^{\prime}(\vt{r}) = \int_0^{+\infty} w(t)c(\vt{r}(t), \vt{v}, \vt{f})\mathrm{d} t\,,
\end{equation}
where $w(t)$ is an unbiased and occlusion-aware weight function used in~\cite{wang2021neus}. The color net outputs a three-channel color vector, to supervise it using gray-scale images, we
% To utilize the color information available in a fraction of the photos, suppose that the corresponding color of the ray $\vt{r}$ is $\mat{C}(\vt{r})$ from a gray-scale image $\I$
use perceptual weights~\cite{wang2011information} to convert the output color to gray-scale value, \ie, for $\mat{C}^{\prime}(\vt{r}) = (c_r, c_g, c_b)$, we propose the function $g:\mathbb{R}^3 \to \mathbb{R}$ and
\begin{equation}
    g(\mat{C}^{\prime}(\vt{r})) = w_r c_r + w_g c_g + w_b c_b\,,
\end{equation}
where $W_r=0.2126$, $W_g=0.7152$ and $W_b=0.0722$. 
The loss for ray color $\mat{C}^{\prime}(\vt{r})$ in image with true color $\mat{C}(\vt{r})$ is
\begin{equation}
    \mathit{l}_c(\vt{r}) = \begin{cases}
    \frac{1}{2}  |\mat{C}(\vt{r}) - g(\mat{C}^{\prime}(\vt{r}))|^2, & \text{if}~ \vt{r}~\text{is gray-scale} \,, \\
    \frac{1}{2}  |\mat{C}(\vt{r}) - \mat{C}^{\prime}(\vt{r})|^2,              & \text{otherwise}\,.
\end{cases}
\end{equation}
With the color appearance embedding loss, we weakly supervise on gray-scale images and strongly on color images. 
% \lu{why low/high frequency matters?} \david{becaues we have few color images and color is relatively smooth (low frequency) therefore weak supervision is enough}
% This works because color is low frequency information (this is used in many compression algorithms) which is weakly supervised by grayscale images (intensity) and strongly by color images. In other words, the network can disambiguate the colors based on few examples. 
% The high frequency derails are supervised by both grayscale and color images without extra effort??
% The total loss is hence
% \begin{equation}
%     \mathit{l}(\vt{r}) = \mathit{l}_c(\vt{r}) + \lambda \mathit{l}_g(\vt{r})\,,
% \end{equation}
% where $\lambda$ is a hyperparameter.
% \lu{how to use x in geometry loss and ray together is unclear}
%-------------------------------------------------------------------------

\section{Experiments}\label{sec:eva}
% We remove blobs covering the buildings from evaluation views

\begin{figure}[h]
\centering
\setlength{\tabcolsep}{2pt}
\newcolumntype{Y}{>{\centering\arraybackslash}p{0.3\linewidth}}
\begin{tabular}{YYY}
St. Michael Church & Hotel International & Observatory \\
\includegraphics[width=\linewidth]{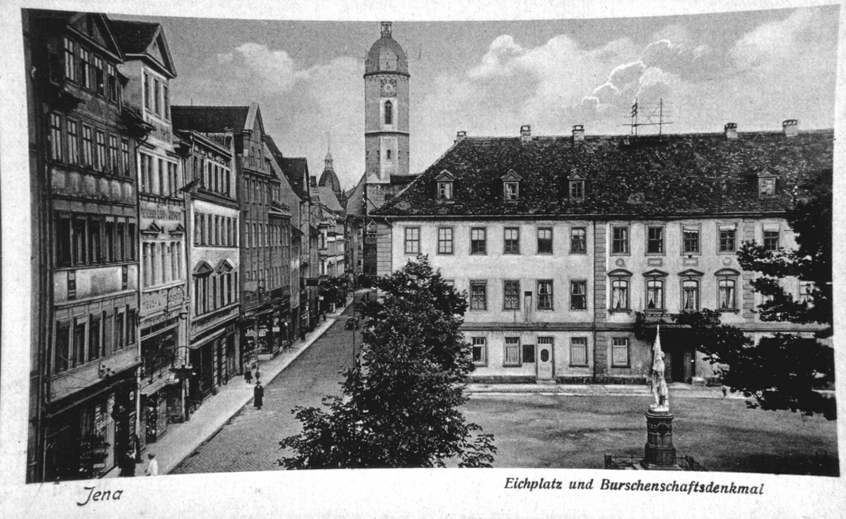} &
\includegraphics[width=\linewidth]{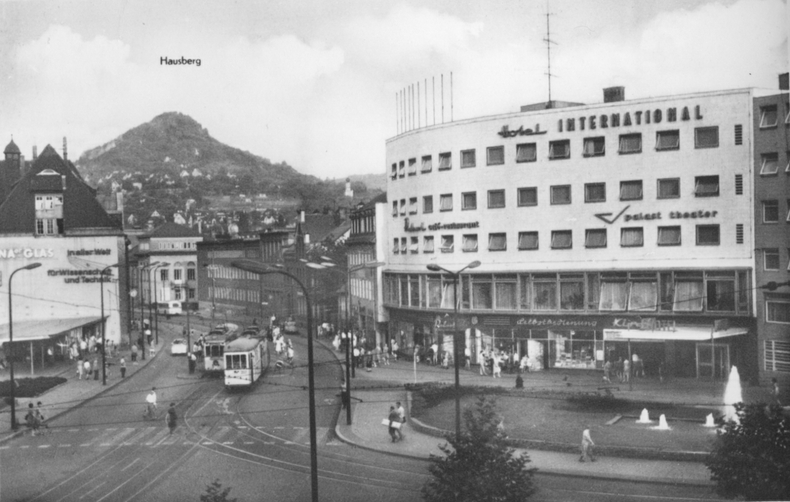} &
\includegraphics[width=\linewidth]{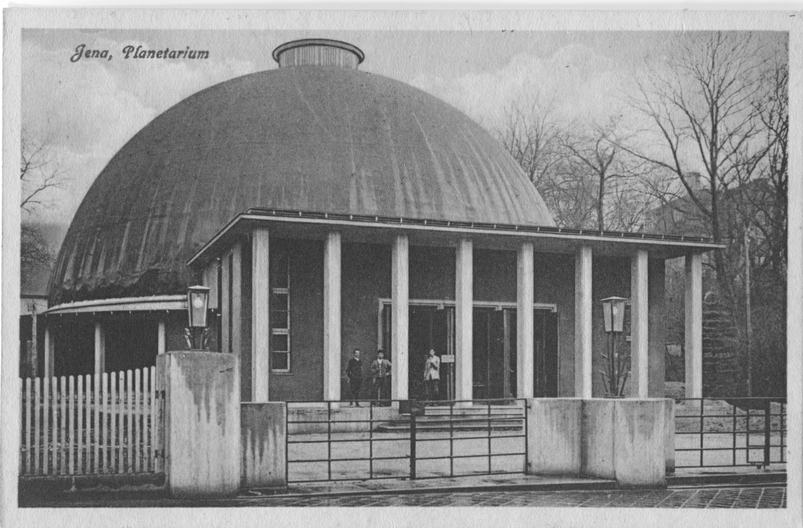}\\
\includegraphics[width=\linewidth]{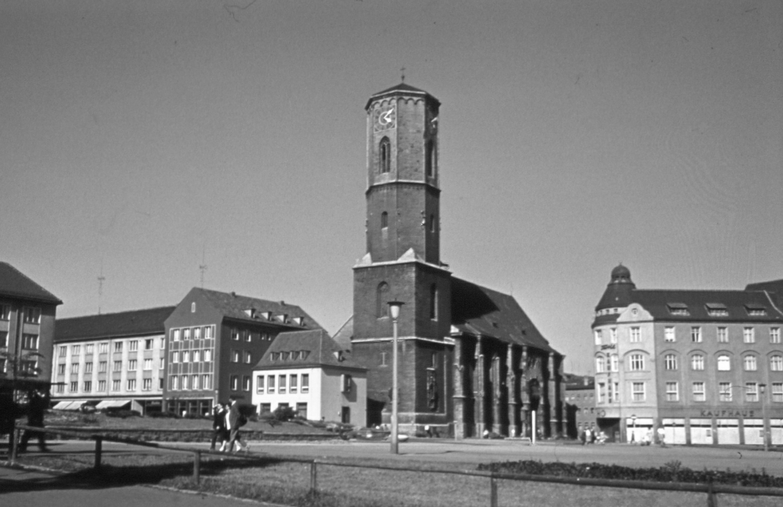} &
\includegraphics[width=\linewidth]{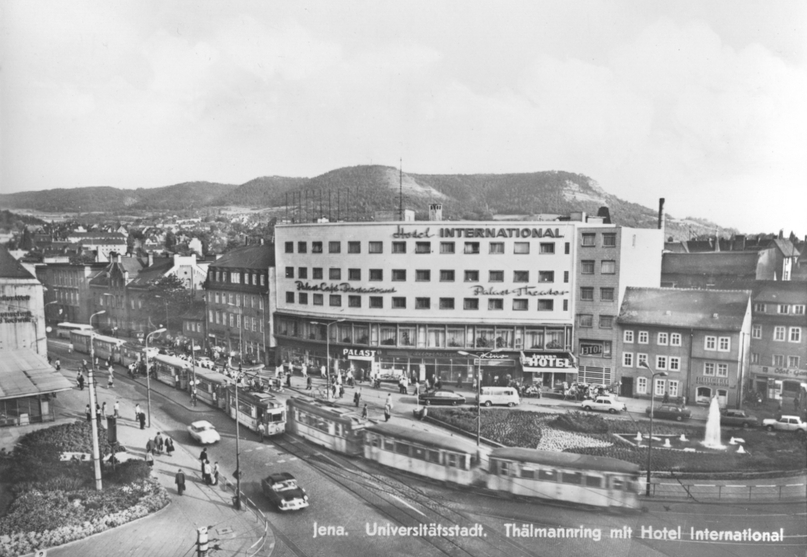} &
\includegraphics[width=\linewidth]{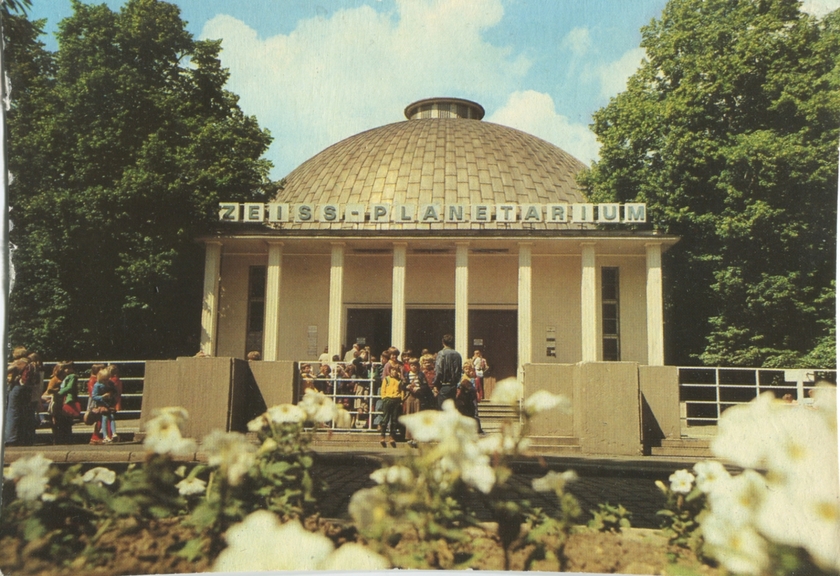}
\end{tabular}
\caption{Example images from the historical datasets~\cite{jena} The lighting and view angles are largely changed in all datasets. Images are far from targeted buildings as well. All the features make reconstruction challenging. }\label{fig:historical_dataset_img}
\vspace{-0.3cm}
\end{figure}

\paragraph{Validation Dataset} We evaluate our method on the \textbf{historical datasets} listed in~\cref{tab:dataset} and one \textbf{modern dataset} which shows the Brandenburg Gate~\cite{brandenburg}. 
We exclude certain images in the historical dataset where the main object is not visible or the building is demolished. The total number of images used for training is listed in~\cref{tab:dataset}. The historical datasets consist of significantly fewer images compared to the Brandenburg Gate dataset. They also introduce challenges such as limited viewing angles (the Observatory dataset), or a wide range of lighting conditions (the Hotel dataset, see in~\cref{fig:historical_dataset_img}).
% The Observatory is a challenging scene due to the limited viewing angle concentrated around the frontal view. The uniform color of the building makes it extremely difficult to reconstruct, especially the dome and the wall behind the columns.
% The St Michael Church has the fewest images in the dataset, but it has a wide range of viewing angles including aerial photos. The subject is a complex square with multiple houses and a church.
% The former Hotel International building has a simple geometry and a good distribution of viewing angles, but is limited by the number of images and the wide range of lighting conditions, containing multiple night photos with neon lights illuminating the building.
The Brandenburg Gate dataset provides corresponding ground truth in the form of LiDAR measurements~\cite{sun2022neuconw}. Hence, we use it for quantitative evaluation. 
To close the gap between the modern and historic domains, we transform $90\%$ of the images in the Brandenburg Gate dataset into gray-scale. To show the influence of input data quantity, we sample $10$ and $90$ images from the dataset.% and we also transform $90\%$ of these images to gray-scale for testing the color appearance embedding introduced in~\cref{subsec:color}. 
% For these numbers are close to our historic datsets
\vspace{-0.4cm}
\paragraph{Implementation Details} For our method and its comparison to other methods we build upon the implementation of SDFStudio~\cite{Yu2022SDFStudio}. For dense point cloud generation, we import the camera poses and feature matches obtained by COLMAP~\cite{colmap} and a combination of state-of-the-art keypoint detector and feature matching algorithms \cite{tyszkiewicz2020disk,detone18superpoint,lindenberger2023lightglue} via the bundler format and apply the segmentation masks. 
We use 8 layers with 512 hidden units for the geometry MLP and 4 layers
with 256 hidden units for the color MLP. 

For the historic datasets, we select a sampling radius $V_{sfm}$ roughly 2 times the radius of the encapsulating sphere of the main object or building of interest. 
For the geometry loss \cref{eq:geometry_loss} we set $\lambda=0.1$. The voxel grid used for accelerated sampling is updated every 5k iterations.
We sample the color network~\cref{eq:color_net} at the vertices and save it as vertex color. During inference, we use the average appearance embedding vector for the color network. We remove floating blobs occluding the view from the validation views disconnected from the object of interest for historic datasets.
We run all experiments on 4 NVIDIA A100 GPUs for $100.000$ iterations with a batch size of 2048 per GPU.
For the final output mesh, we only extract a mesh within the $V_{sfm}$ radius using Marching Cubes~\cite{marching_cubes} algorithm with a grid resolution of 1024. 

% \david{we sample an image and use the subset of dense pcd that are visible from this imaeg.}

% \david{the teaser uses 16k batch_size}\lu{mention in supp please}

\subsection{Surface Reconstruction Results}
\begin{figure*}[!t]
\centering
\setlength{\tabcolsep}{2pt}
\newcolumntype{Y}{>{\centering\arraybackslash}p{0.18\textwidth}}
\begin{tabular}{lYYYYY}
    & Metashape~\cite{metashape2019} & NeusW~\cite{sun2022neuconw} & Neus-facto~\cite{Yu2022SDFStudio} & Ours & Ours color\\
    \multirow{2}{*}{\rotatebox{90}{Theater}} & 
    \includegraphics[width=\linewidth]{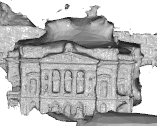} &
    \includegraphics[width=\linewidth, clip, viewport=250 120 600 404]{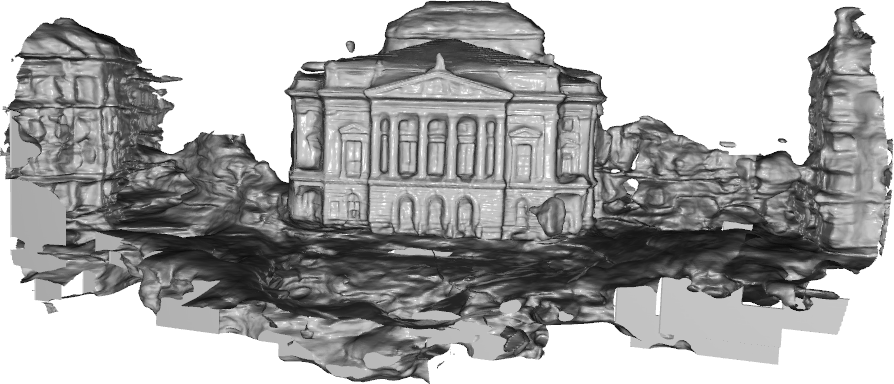} & 
    \includegraphics[width=\linewidth, clip, viewport=250 120 600 404]{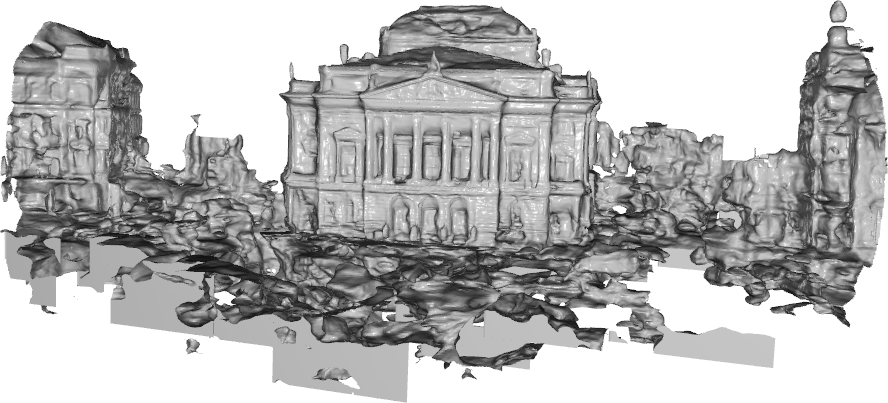} & 
    \includegraphics[width=\linewidth, clip, viewport=250 120 600 404]{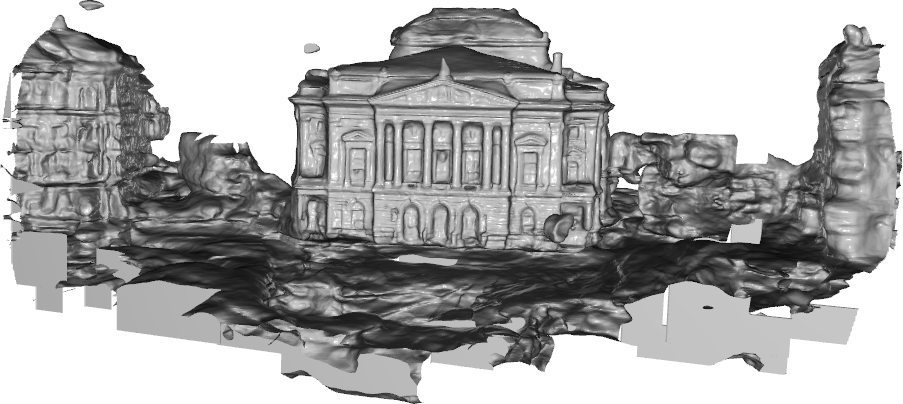} &
    \includegraphics[width=\linewidth, clip, viewport=250 120 600 404]{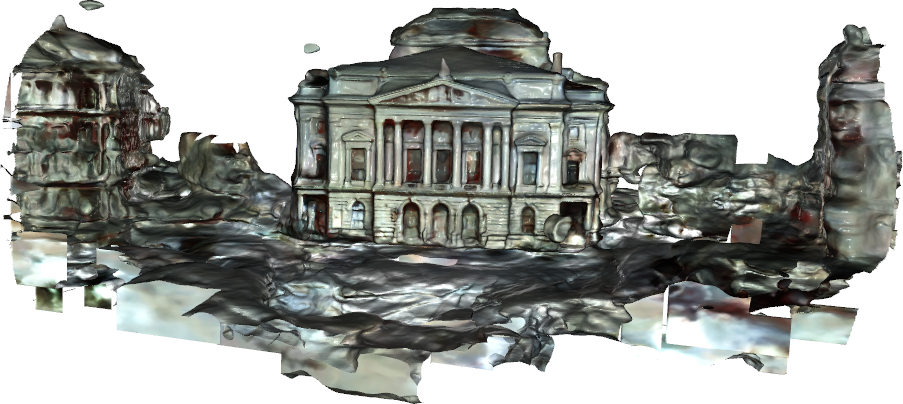} \\
    & \includegraphics[width=\linewidth]{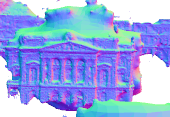} &
    \includegraphics[width=\linewidth, clip, viewport=250 120 600 404]{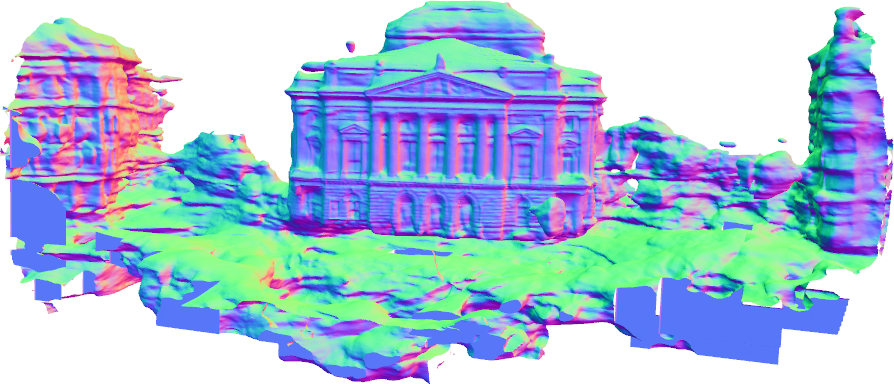} & 
    \includegraphics[width=\linewidth, clip, viewport=250 120 600 404]{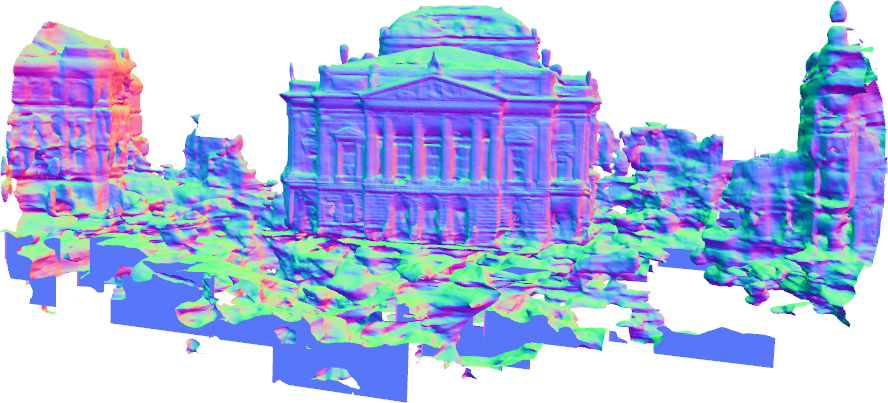} & 
    \includegraphics[width=\linewidth, clip, viewport=250 120 600 404]{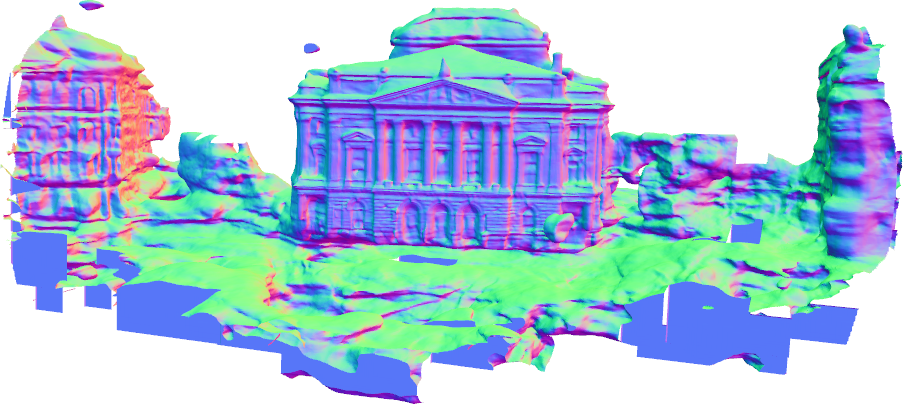} &
    \includegraphics[width=\linewidth, clip, viewport=360 200 500 300]{imgs/comparison/theater/ours_normal.png} \\
    
    \multirow{2}{*}{\rotatebox{90}{Observatory}} & 
    \includegraphics[width=\linewidth]{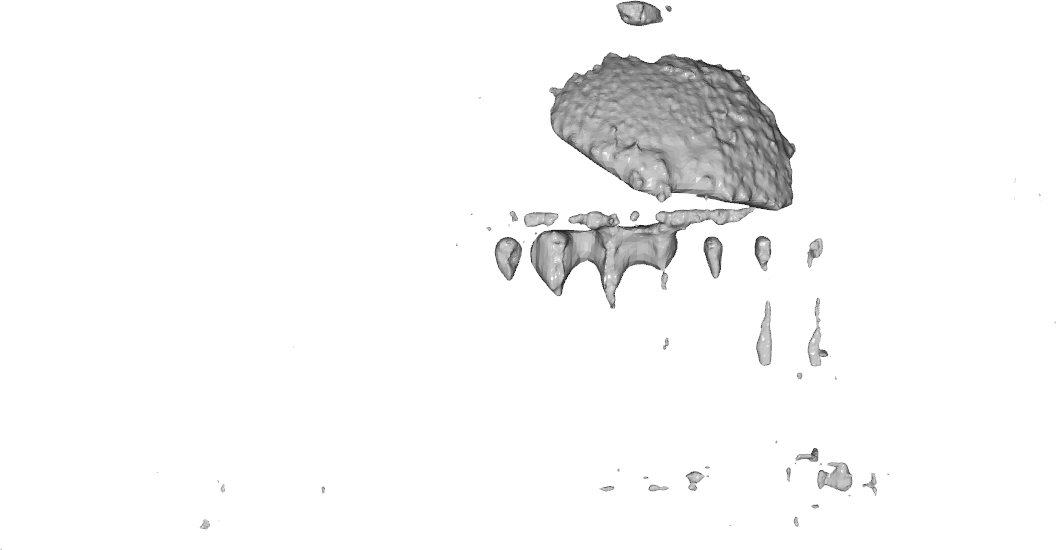} & 
    \includegraphics[width=\linewidth, clip, viewport=100 0 800 452]{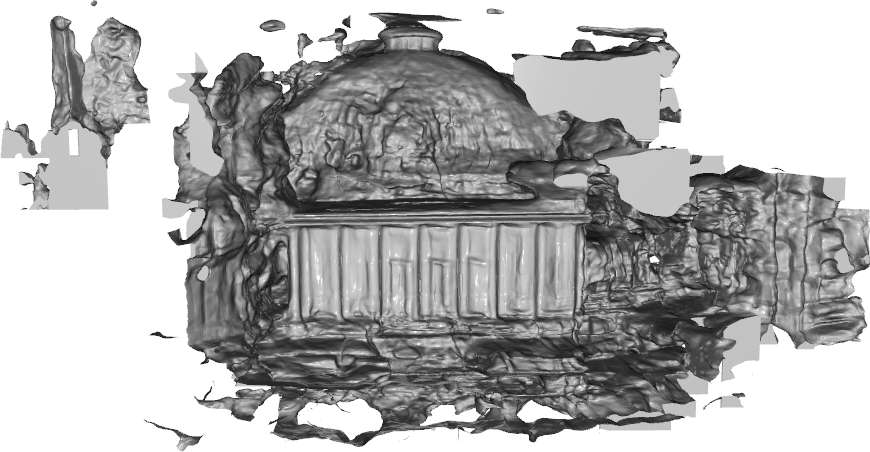} & 
    \includegraphics[width=\linewidth, clip, viewport=200 100 800 524]{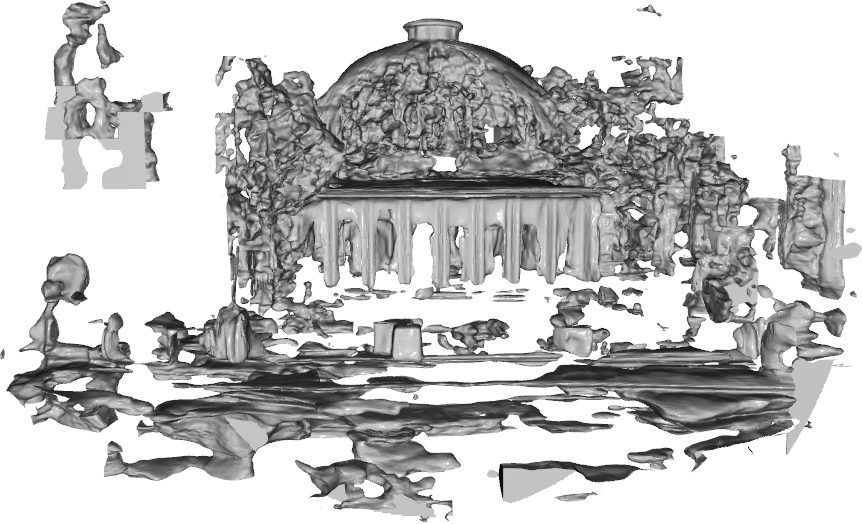} &
    \includegraphics[width=\linewidth, clip, viewport=100 200 800 707]{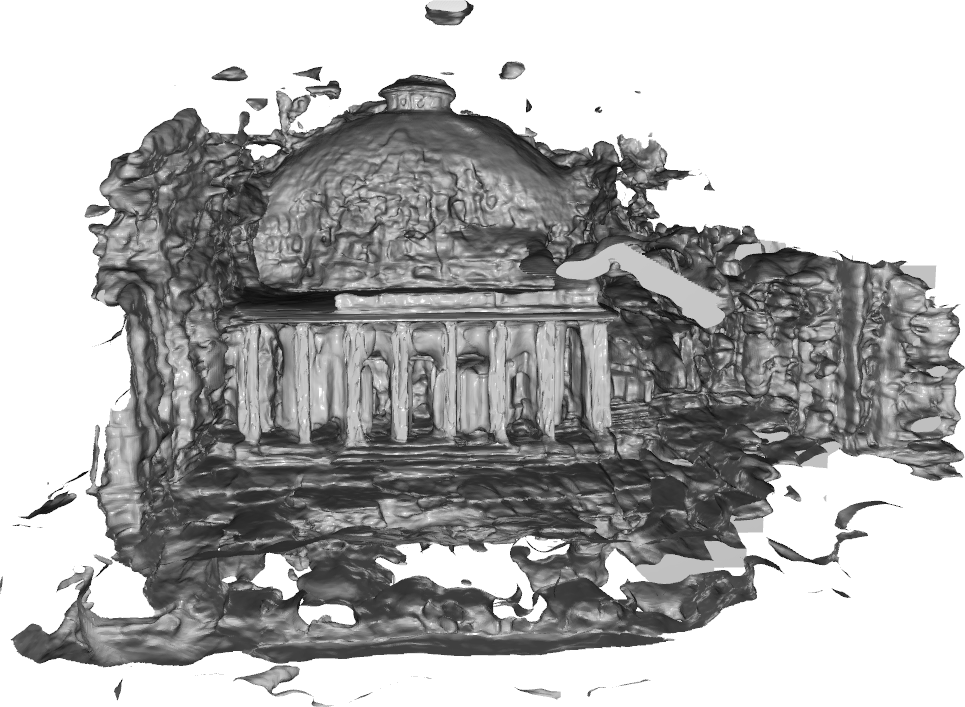} &
    \includegraphics[width=\linewidth, clip, viewport=100 200 800 707]{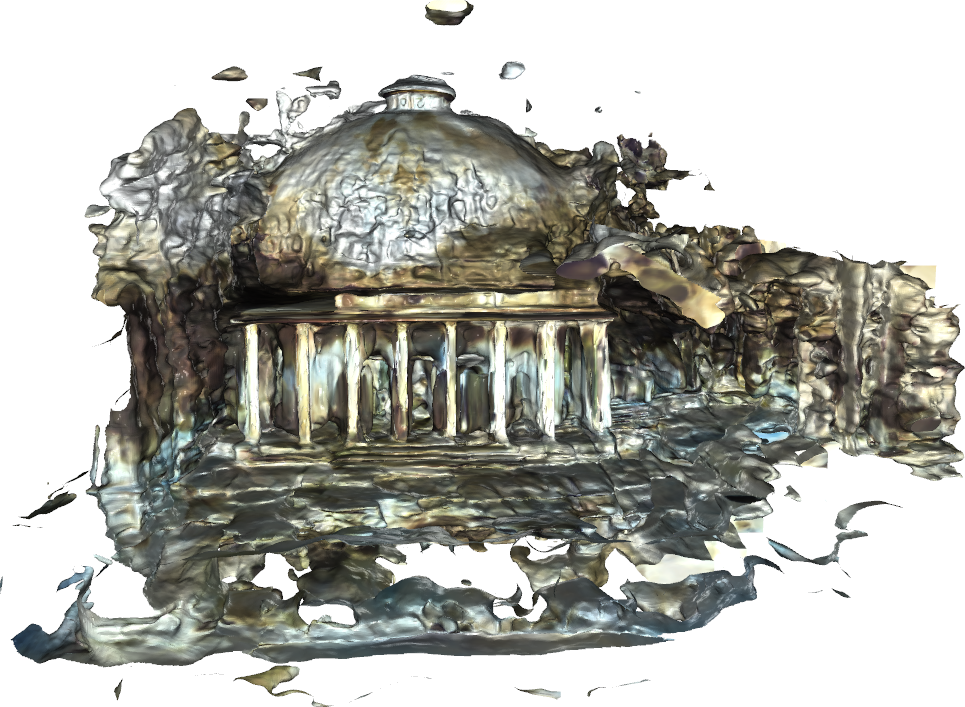} \\
   & \includegraphics[width=\linewidth]{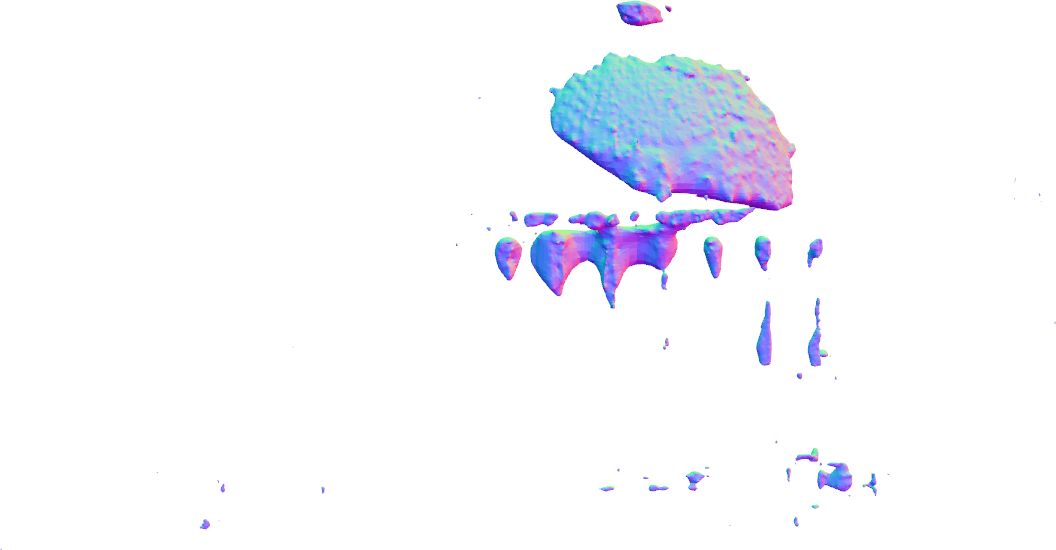} & 
    \includegraphics[width=\linewidth, clip, viewport=100 0 800 452]{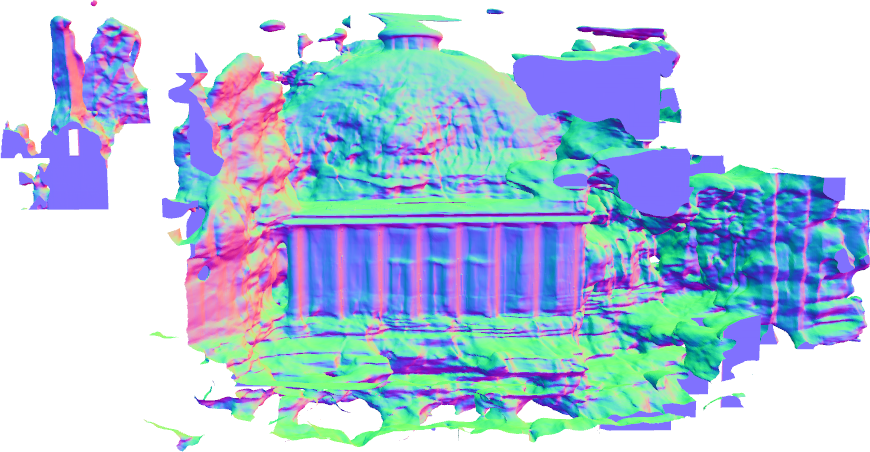} & 
    \includegraphics[width=\linewidth, clip, viewport=200 100 800 524]{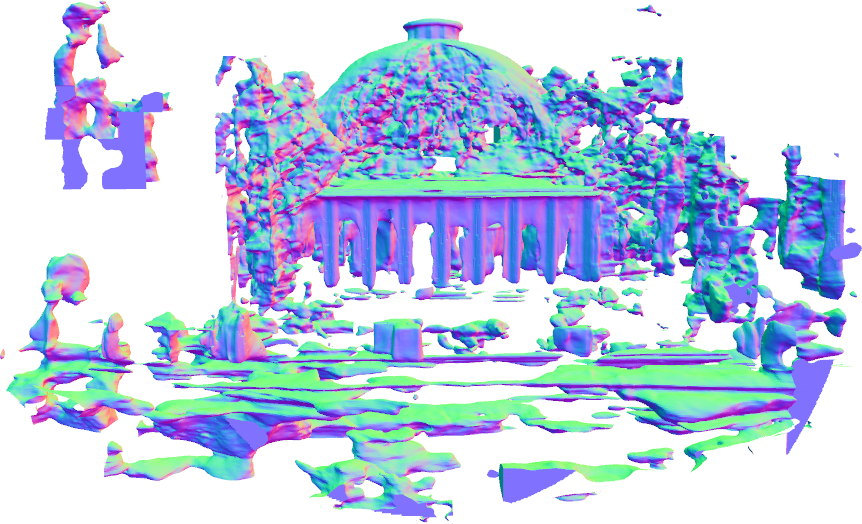} &
    \includegraphics[width=\linewidth, clip, viewport=100 200 800 707]{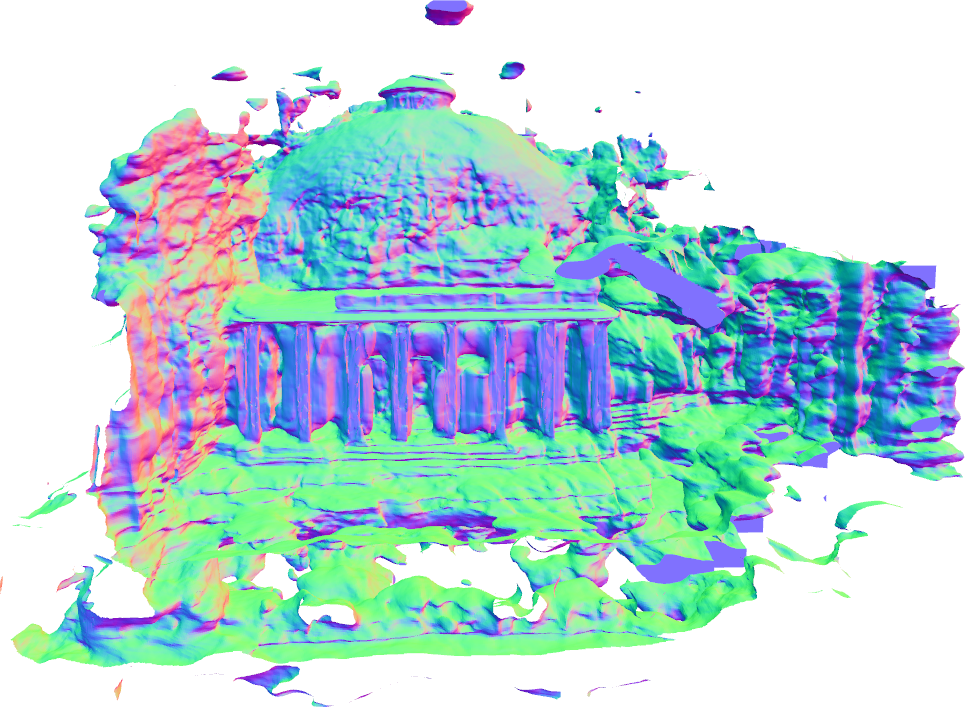} &
    \includegraphics[width=\linewidth, clip, viewport=200 250 600 450]{imgs/comparison/observatory/ours_normal_crop.png}\\
    
    \multirow{2}{*}{\rotatebox{90}{ St. Michael Church}} & 
    \includegraphics[width=\linewidth]{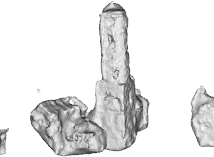} & 
    \includegraphics[width=\linewidth, clip, viewport=160 205 610 640]{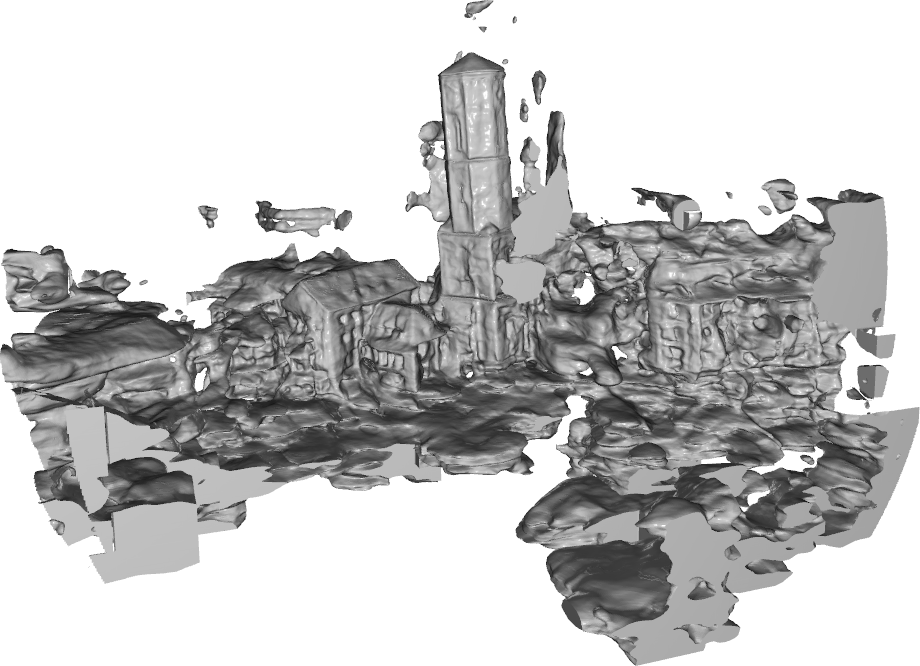} & 
    \includegraphics[width=\linewidth, clip, viewport=160 205 610 640]{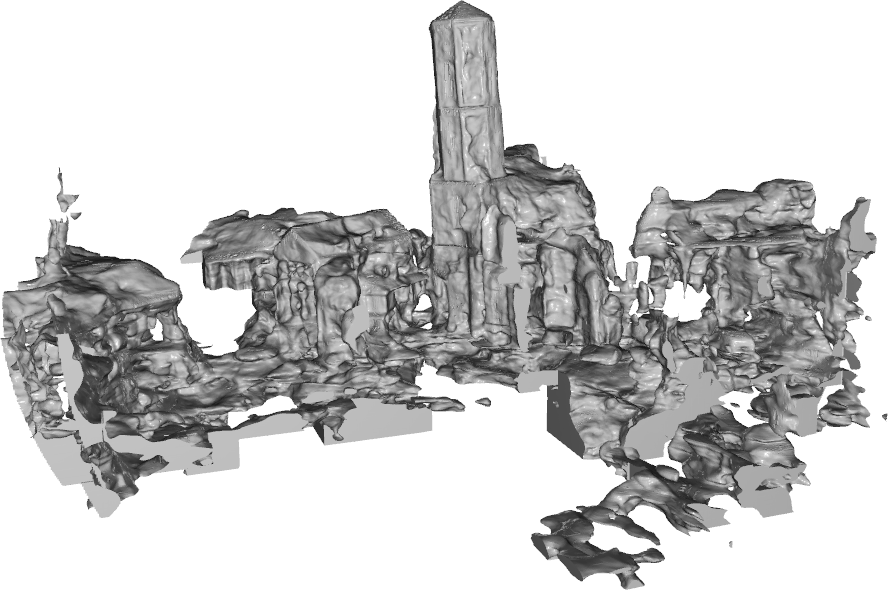} &
    \includegraphics[width=\linewidth, clip, viewport=160 205 610 640]{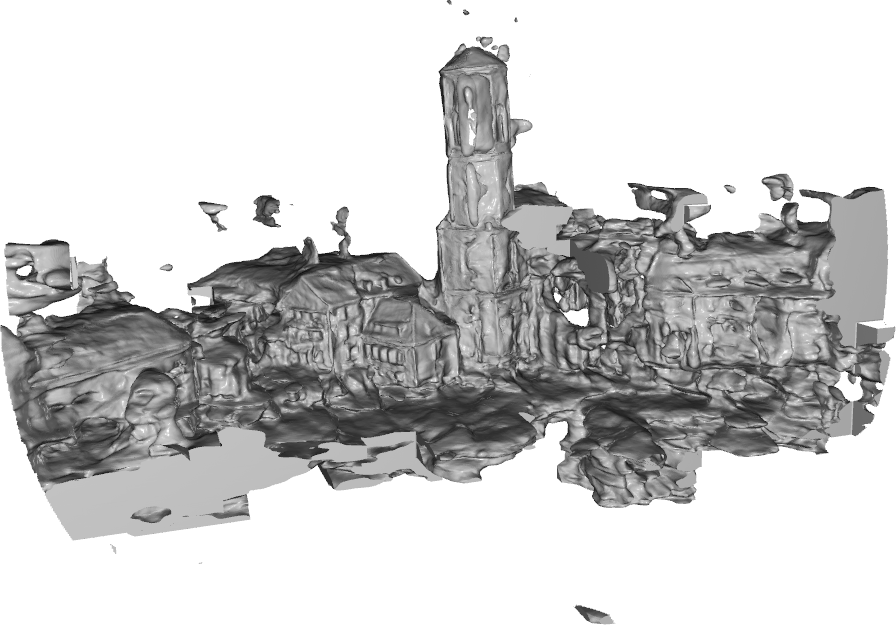} &
    \includegraphics[width=\linewidth, clip, viewport=160 205 610 640]{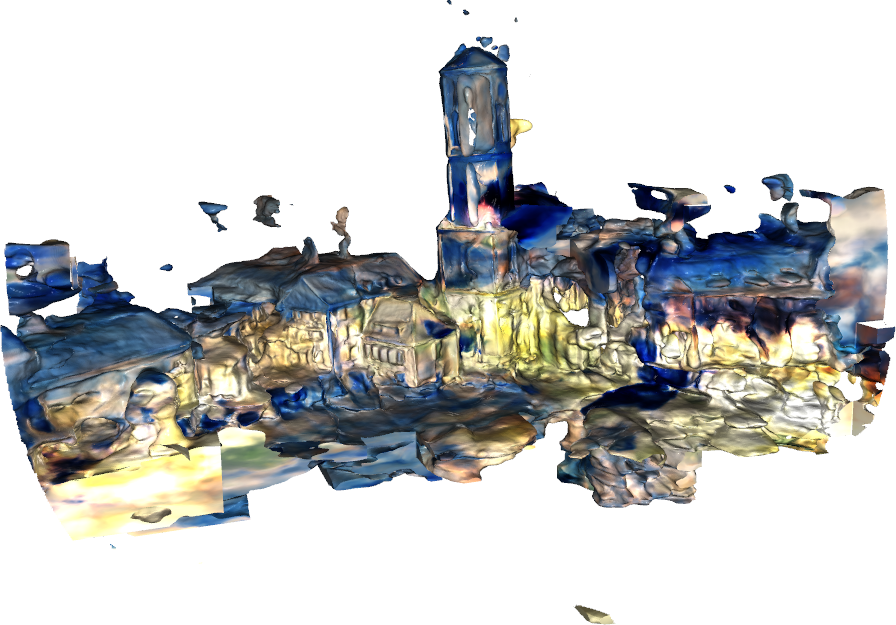}\\
    & \includegraphics[width=\linewidth]{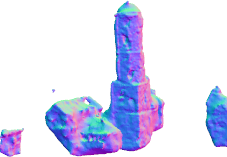} & 
    \includegraphics[width=\linewidth, clip, viewport=160 205 610 640]{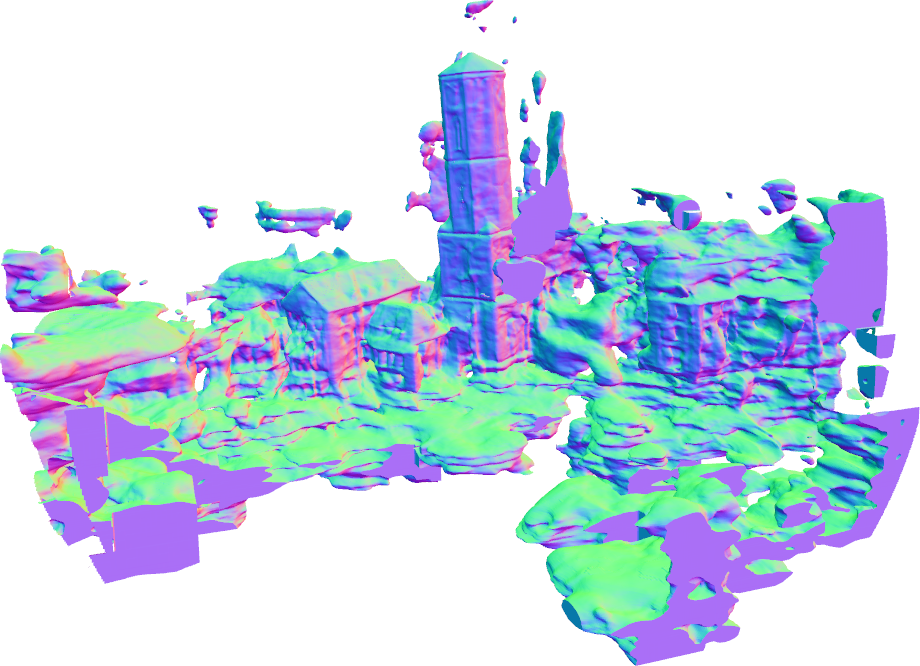} & 
    \includegraphics[width=\linewidth, clip, viewport=160 205 610 640]{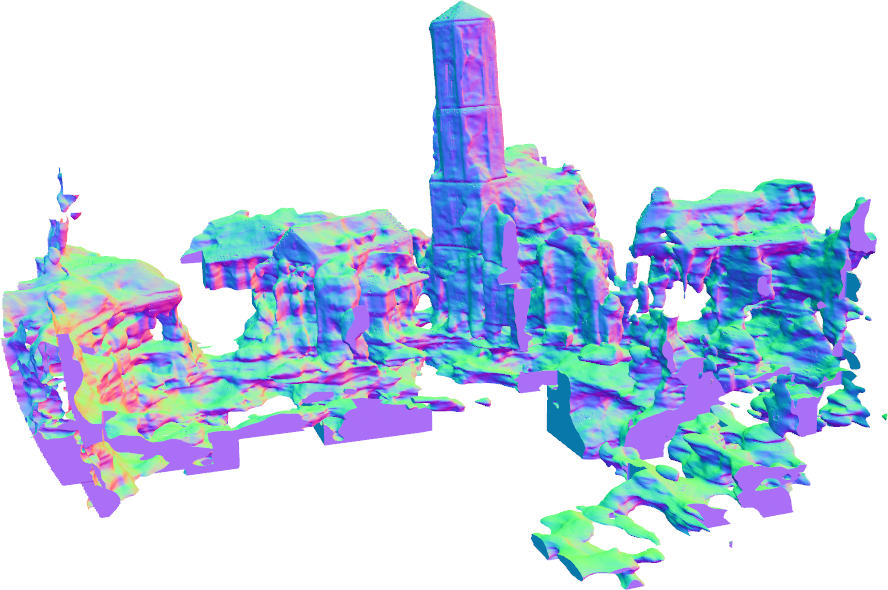} &
    \includegraphics[width=\linewidth, clip, viewport=160 205 610 640]{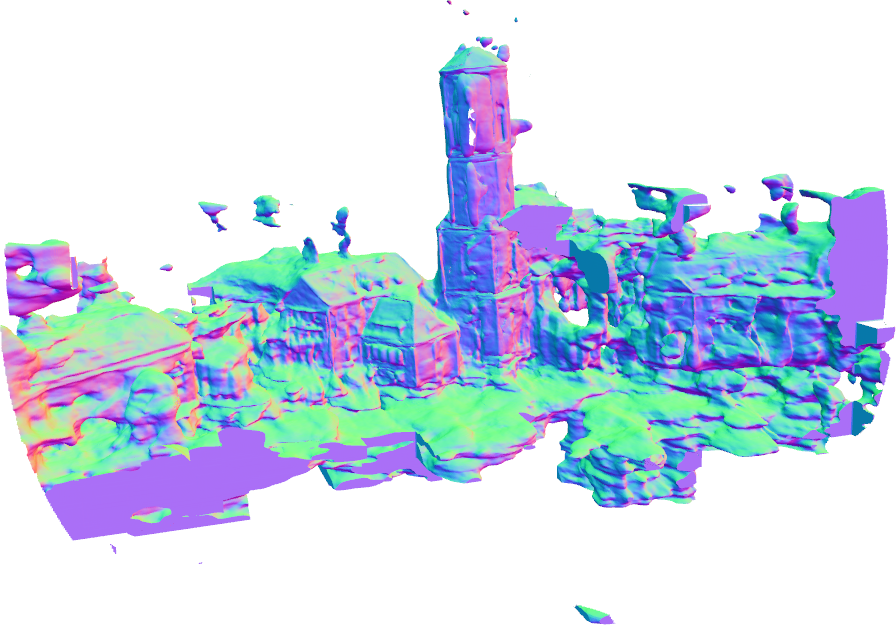} & 
    \includegraphics[width=\linewidth, clip, viewport=300 205 510 400]{imgs/comparison/church/ours_normal_crop.png} \\
    
    \multirow{2}{*}{\rotatebox{90}{\footnotesize Hotel International}} & 
    \includegraphics[width=\linewidth, clip, viewport=100 0 750 300]{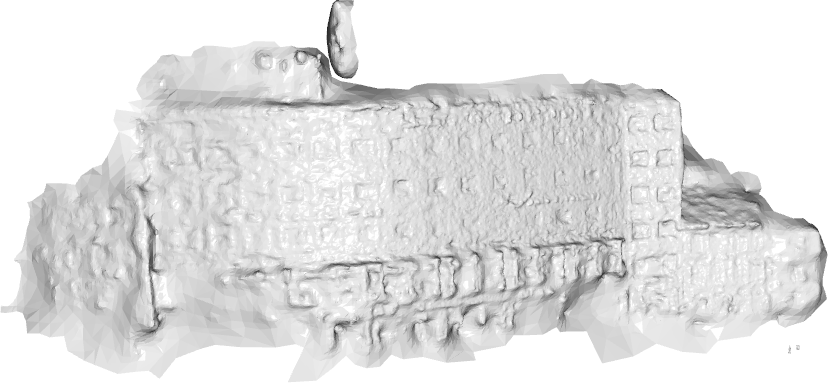} & 
     \includegraphics[width=\linewidth, clip, viewport=150 0 750 300]{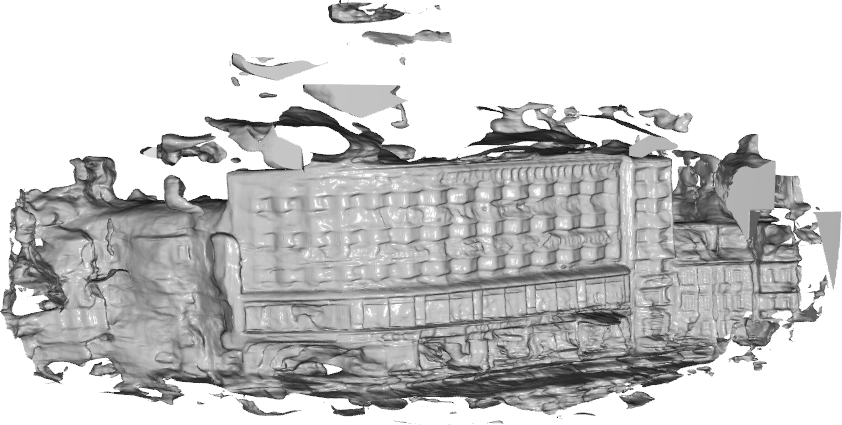} & 
     \includegraphics[width=\linewidth, clip, viewport=150 0 750 300]{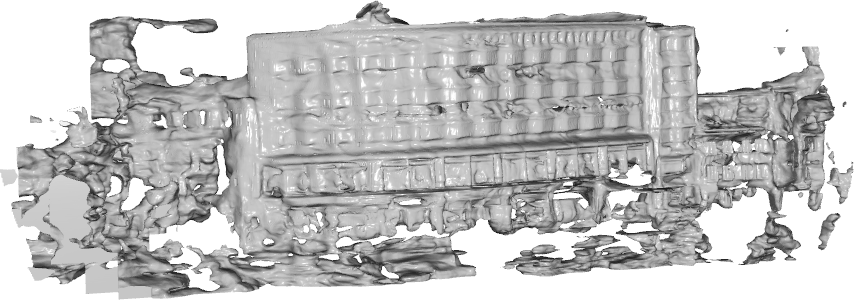} &
     \includegraphics[width=\linewidth, clip, viewport=150 0 750 300]{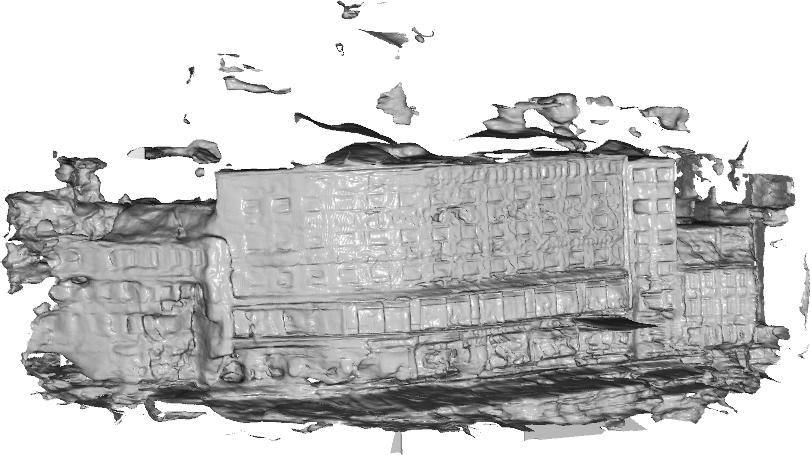} &
     \includegraphics[width=\linewidth, clip, viewport=150 0 750 300]{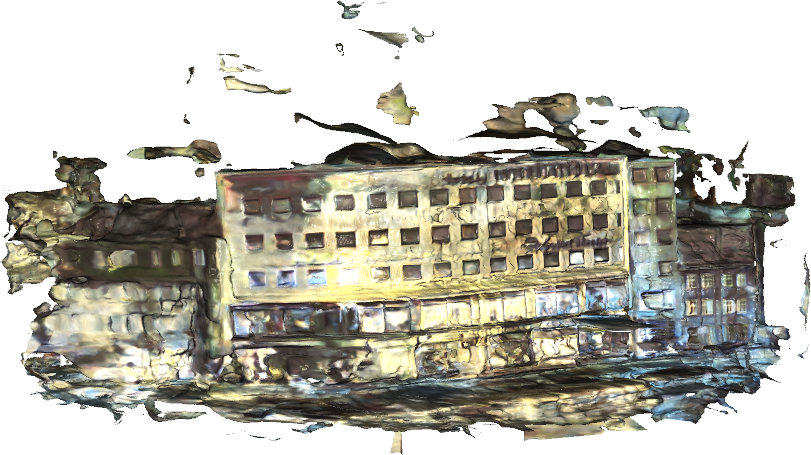}\\
     & \includegraphics[width=\linewidth, clip, viewport=100 0 750 300]{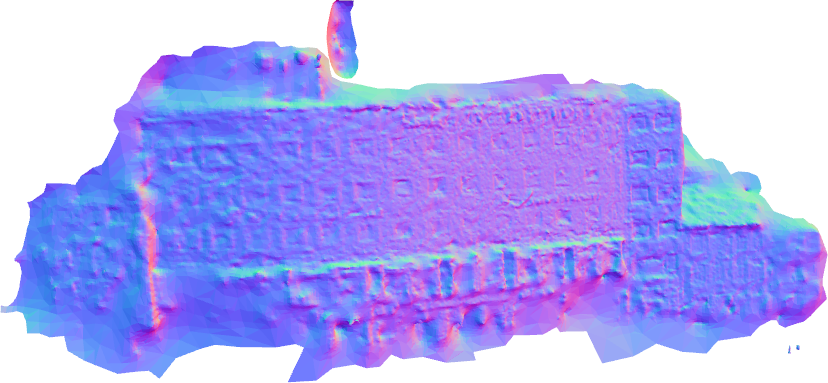} & 
     \includegraphics[width=\linewidth, clip, viewport=150 0 750 300]{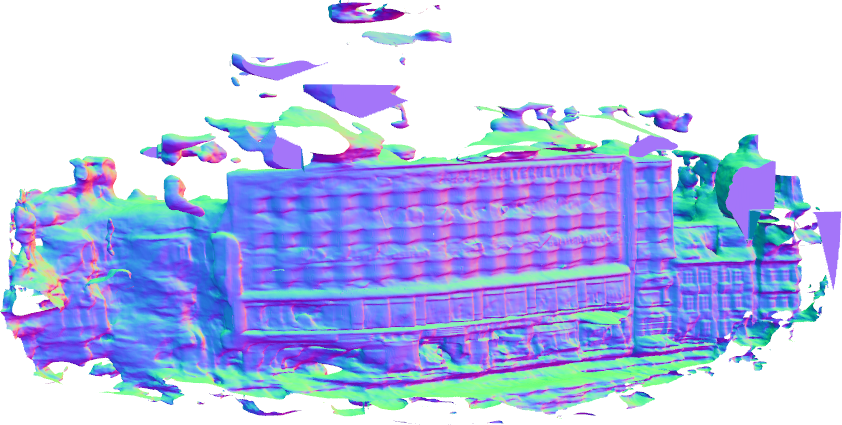} & 
     \includegraphics[width=\linewidth, clip, viewport=150 0 750 300]{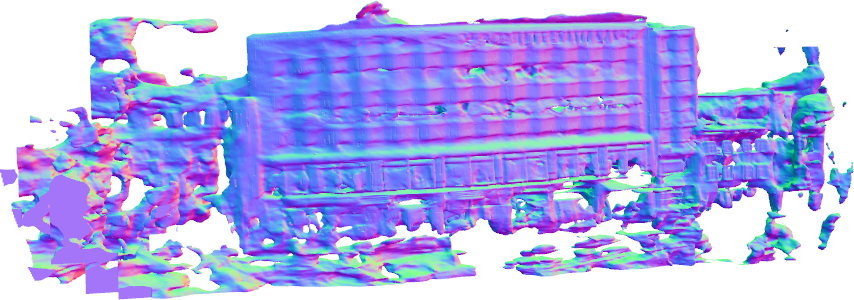} &
     \includegraphics[width=\linewidth, clip, viewport=150 0 750 300]{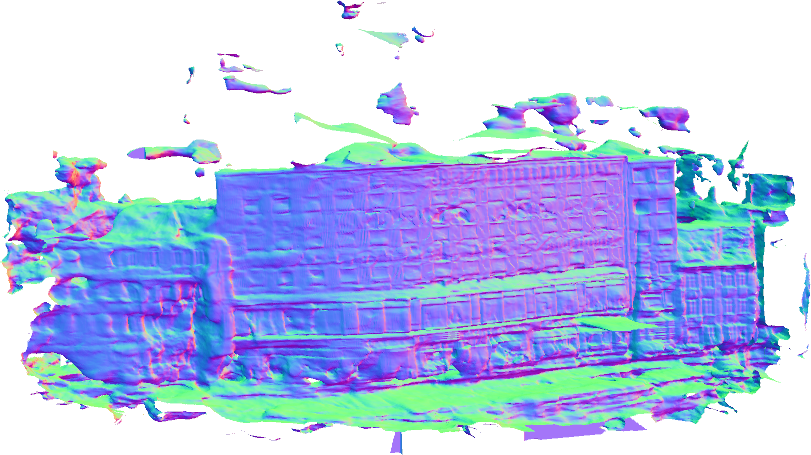} & \includegraphics[width=\linewidth, clip, viewport=250 100 450 200]{imgs/comparison/hotel/ours_cut_normal_crop.png} \\
    %  & \includegraphics[width=\linewidth, clip, viewport=100 0 750 300]{imgs/comparison/hotel/metashape_normal_crop.png} & 
    %  \includegraphics[width=\linewidth, clip, viewport=150 0 750 300]{imgs/comparison/hotel/neusW_normal_crop.png} & 
    %  \includegraphics[width=\linewidth, clip, viewport=150 0 750 300]{imgs/comparison/hotel/facto_normal_crop.png} &
    %  \includegraphics[width=\linewidth, clip, viewport=150 0 750 300]{imgs/comparison/hotel/ours_cut_normal_crop.png} &
    %  \includegraphics[width=\linewidth, clip, viewport=150 0 750 300]{imgs/comparison/hotel/ours_cut_normal_crop.png}
      \end{tabular}
  \caption{Reconstructed mesh results compared to other methods. Metashape~\cite{metashape2019} can get clean geometry reconstruction but fails to get the details. Our method is able to provide comparable mesh reconstructions while additionally recovering the color of the mesh. }
  \label{fig:mesh_results}

%   Our method results in a more geometrically accurate models with finer details an retains thin structures. Howerever it also has some artifacts originating from imprecise Sfm point cloud.
%   Color loss causes bigger holes at windows because the color of the interior significantly changes which has a smaller impact in grayscale.
\end{figure*}

In this section, we show our surface reconstruction results in comparison to other methods. ~\cref{fig:mesh_results} shows our reconstructed meshes for four different historical datasets and its comparison to other state-of-the-art conventional~\cite{metashape2019} and learning-based MVS algorithms~\cite{Yu2022SDFStudio, sun2022neuconw} in terms of reconstruction quality. %~\cref{fig:mesh_results} shows the reconstruction visualization results of our methods with compared methods for the historical datasets. 
~\cref{tab:comparison_quantitative} shows the quantitative comparison results on the Brandenburg Gate dataset. 

\begin{figure}[h]
\centering
\setlength{\tabcolsep}{2pt}
\newcolumntype{Y}{>{\centering\arraybackslash}p{0.23\linewidth}}
\begin{tabular}{lYYYY}
& Front right & Normal & Back left & Normal \\
\rotatebox{90}{\footnotesize Metashape} &
\includegraphics[width=\linewidth, clip, viewport=80 50 230 150]{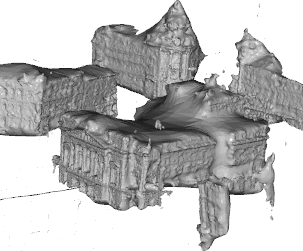} &
\includegraphics[width=\linewidth, clip, viewport=80 50 230 150]{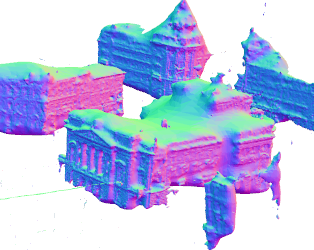} &

\includegraphics[width=\linewidth]{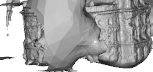} &
\includegraphics[width=\linewidth]{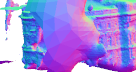} \\

\rotatebox{90}{\footnotesize NeusW} & 
\includegraphics[width=\linewidth, clip, viewport=230 160 600 404]{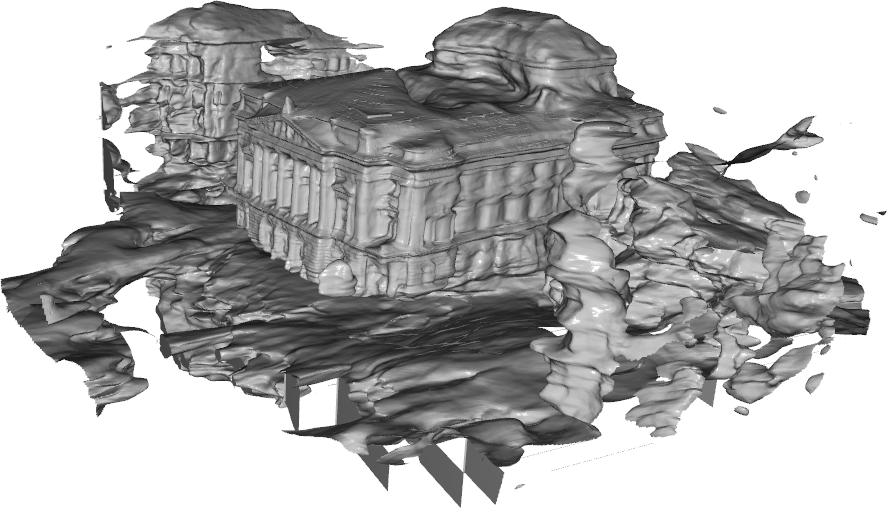} & \includegraphics[width=\linewidth, clip, viewport=230 160 600 404]{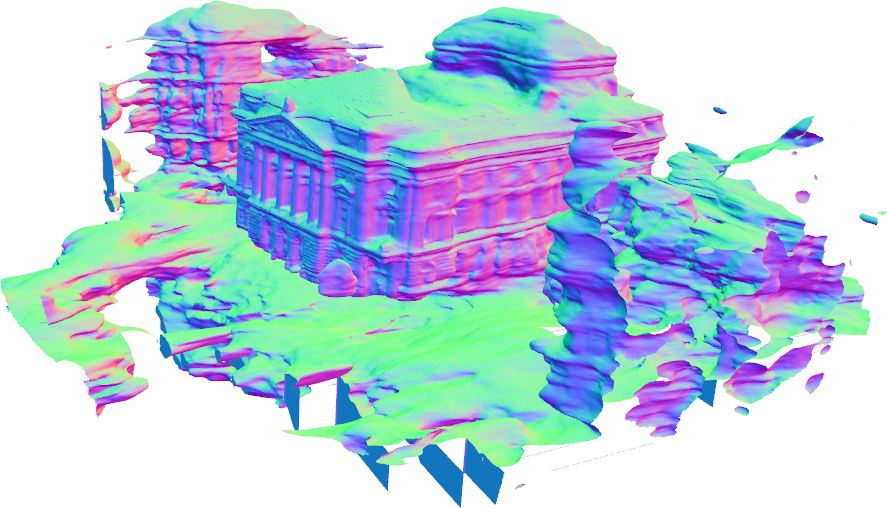} &

\includegraphics[width=\linewidth, clip, viewport=200 150 570 390]{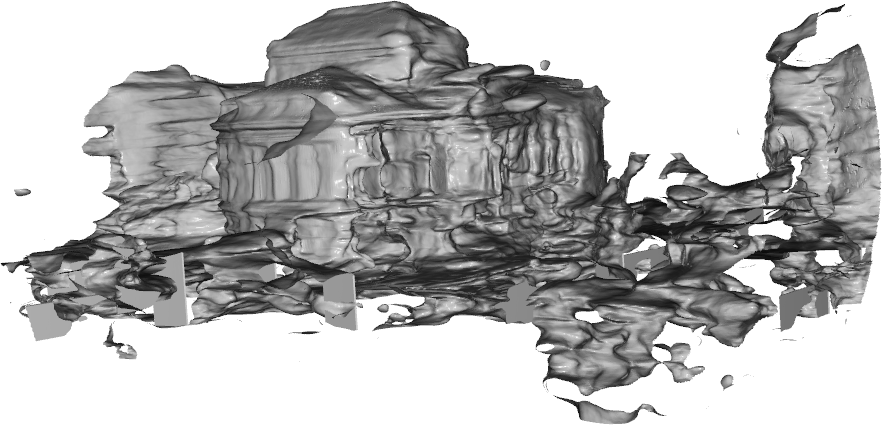} & \includegraphics[width=\linewidth, clip, viewport=200 150 570 390]{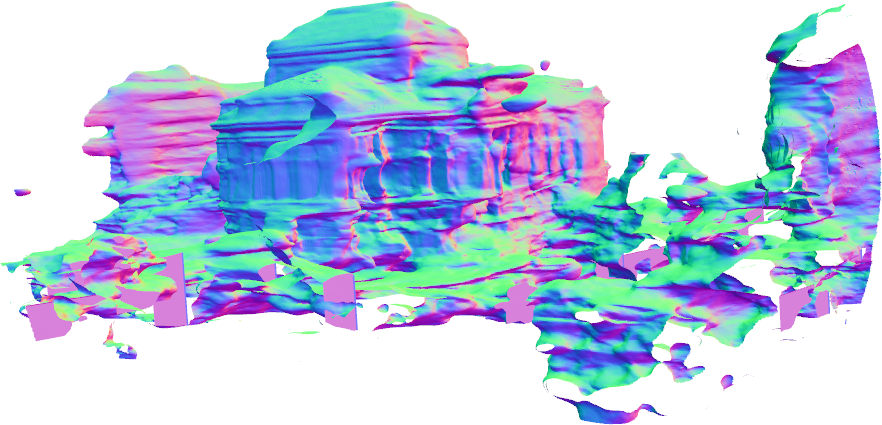} \\

\rotatebox{90}{\footnotesize Neus-facto}&
\includegraphics[width=\linewidth, clip, viewport=230 160 600 404]{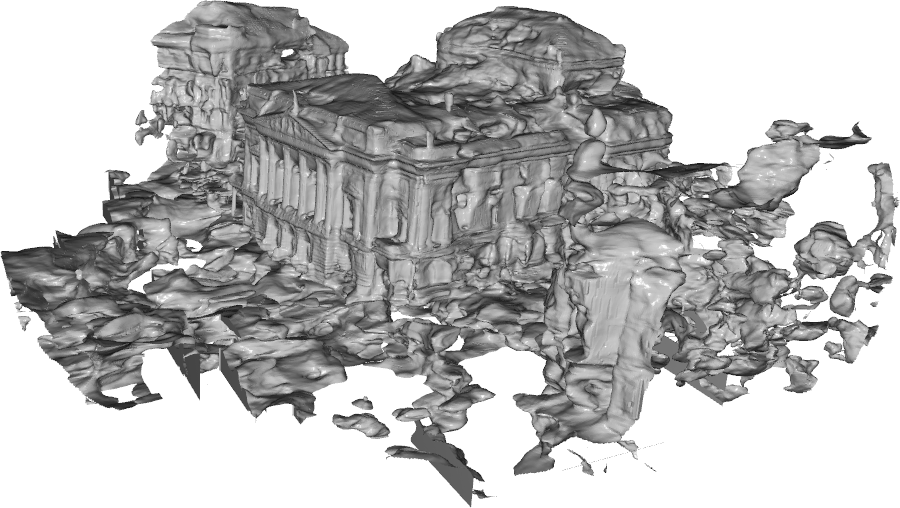} & \includegraphics[width=\linewidth, clip, viewport=230 160 600 404]{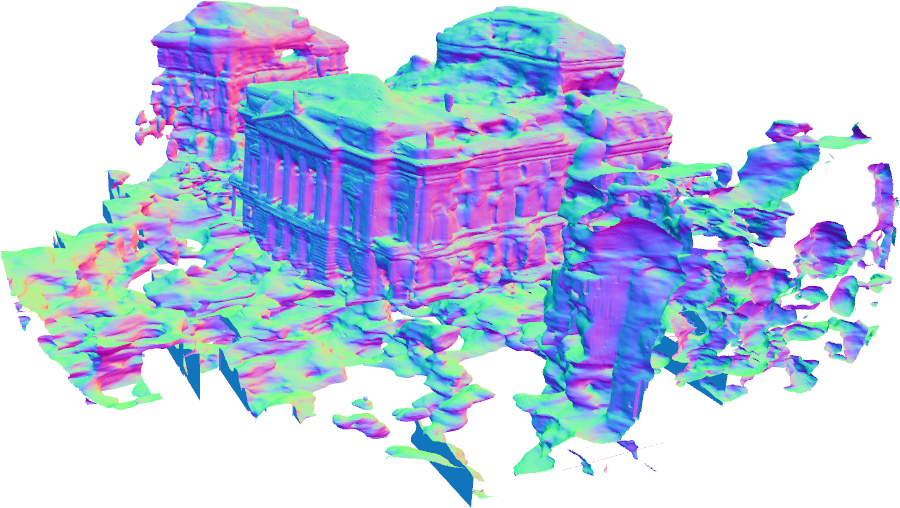} &

\includegraphics[width=\linewidth, clip, viewport=200 150 570 390]{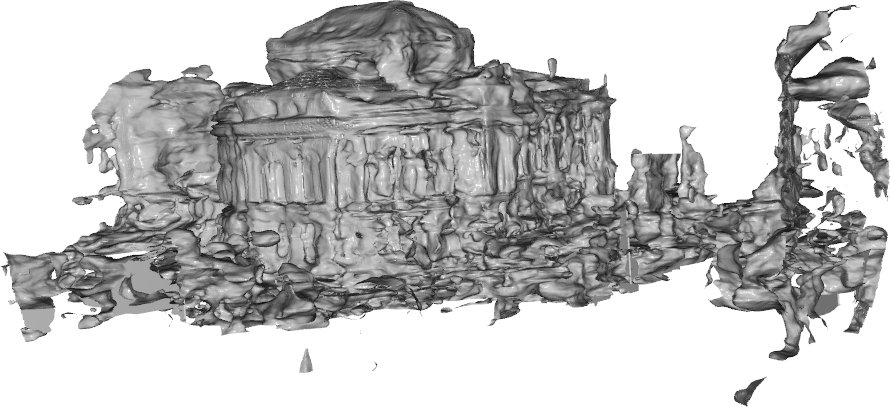} & \includegraphics[width=\linewidth, clip, viewport=200 150 570 390]{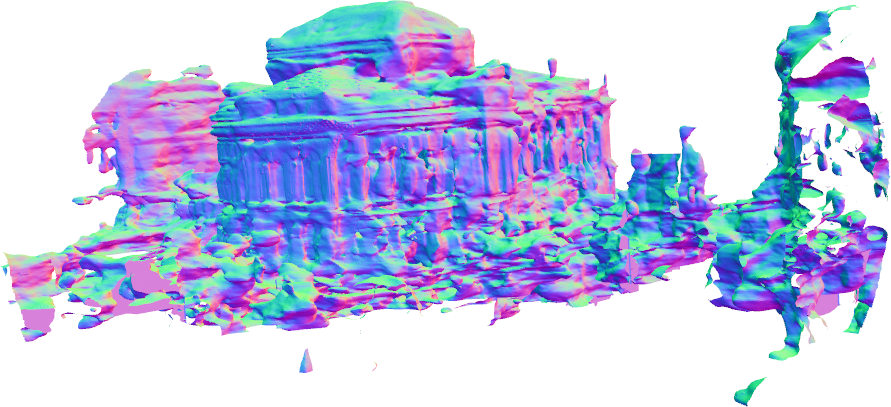} \\

\rotatebox{90}{\small Ours} & 
\includegraphics[width=\linewidth, clip, viewport=230 160 600 404]{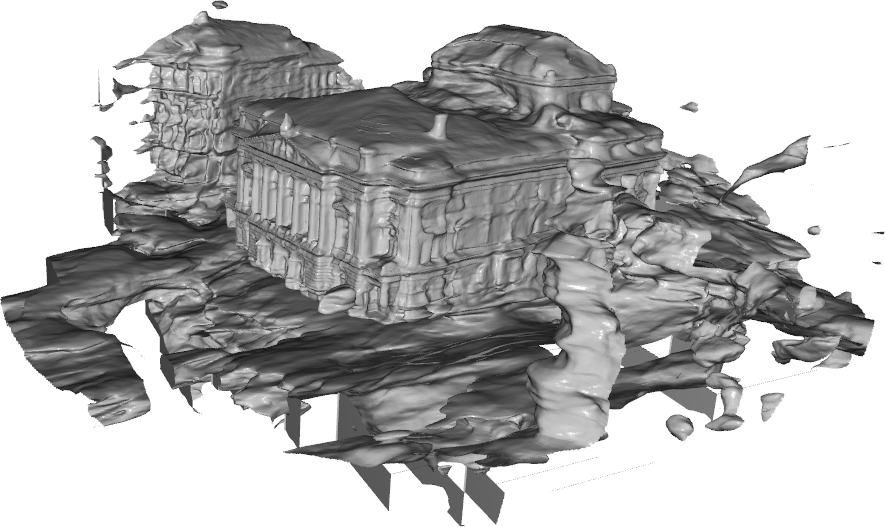} & \includegraphics[width=\linewidth, clip, viewport=230 160 600 404]{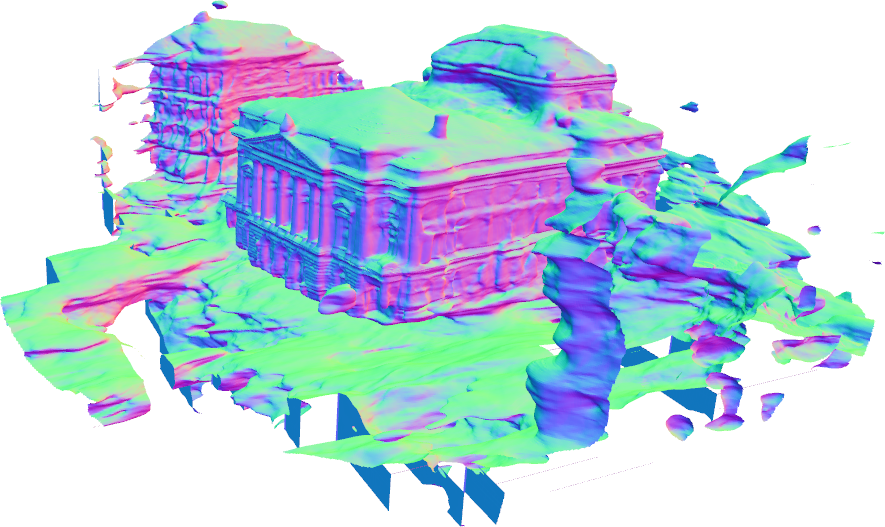} &
\includegraphics[width=\linewidth, clip, viewport=200 150 570 390]{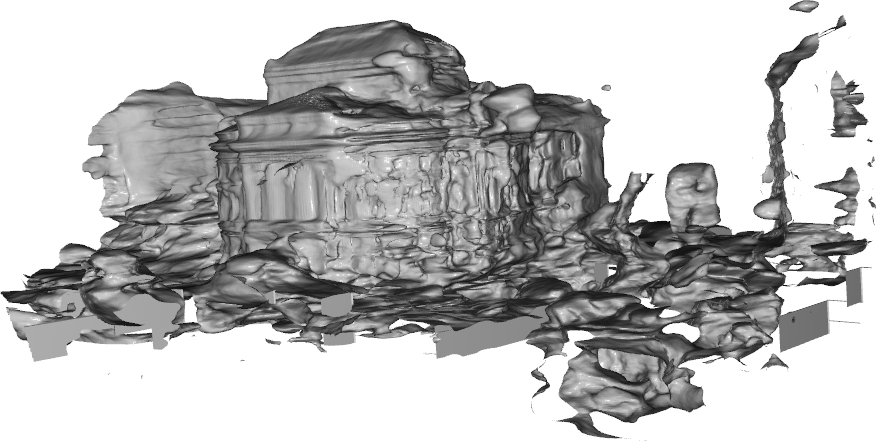} & \includegraphics[width=\linewidth, clip, viewport=200 150 570 390]{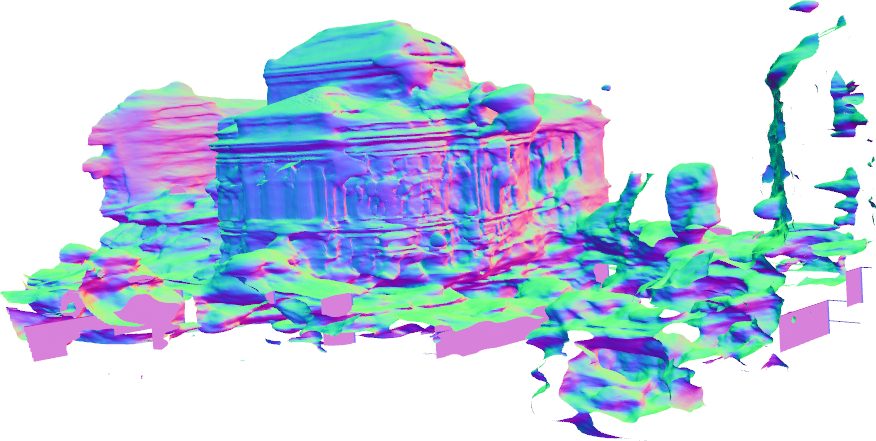}
\end{tabular}
\caption{Reconstruction result of different sides of our dataset showing increasing difficulty in terms of number of images. 
Clearly, the facade can be recovered with intricate detail by all methods, followed by the right side which is satisfactory but smaller deviations arise. The back left corner is the most difficult which is very noisy for almost all methods. Still, ours provides relatively cleaner and complete surface reconstruction.} \label{fig:theater_compare}

\end{figure}

In general, all of the methods achieve the best results on the National Theater dataset, especially for the facade. Qualitatively, our method recovers comparable meshes to other methods. However, as we mentioned in~\cref{sec:dataset}, the back side is more challenging due to the lack of images facing this side. We show the results from different viewing angles in~\cref{fig:theater_compare}. All methods are able to recover the front part of the Theater, but the back part is recovered unsatisfactorily. NeusW~\cite{sun2022neuconw} generates more noise compared to Neus-facto~\cite{Yu2022SDFStudio} and ours. We are able to reconstruct the scene without holes as opposed to Neus-facto~\cite{Yu2022SDFStudio}. For the other datasets, Metashape (using conventional MVS) is only able to recover a small part of the geometry or completely fails, but it tends to provide a closed and clean surface. \par
For the Observatory dataset, in spite of the limited data and challenging setting, learning-based methods can successfully recover the main building with varying degrees of artifacts. Our method gives the most complete and round dome. However, the normal meshes (4th-row in~\cref{fig:mesh_results}) indicate that we are able to recover the pillars correctly while NeusW~\cite{sun2022neuconw} fails on this part. A similar situation happens in the Hotel and Church datasets, ours is able to recover thin structures such as columns and chimneys. We attribute this to the dense point cloud supervision. Finally, our method can recover the colored meshes for the given datasets as shown in the last column in~\cref{fig:mesh_results}.\par
\begin{table*}[!h]
\centering
\begin{tabular}{ll|rrrrrrrrrrrr}
\toprule
 & & \multicolumn{3}{c}{Low} & \multicolumn{3}{c}{Medium} & \multicolumn{3}{c}{High} & \multicolumn{3}{c}{All (AUC)} \\
 \cmidrule{3-14}
& Settings & P & R & F1 & P & R & F1 & P & R & F1 & P & R & F1 \\
\midrule
\multirow{4}{*}{\rotatebox{90}{10 frames}} 
& Metashape & \textbf{73.0} &   8.0 & 14.4 & \textbf{85.0} & 15.4 & 26.0 & \textbf{91.2} & 19.1 & 31.6 & \textbf{88.7} & 22.2 & 35.0 \\
& NeusW        & 37.7 & \underline{26.6} & 31.2 & 55.4 & \textbf{44.0} & \textbf{49.1} & 67.0 & \textbf{55.0} & \textbf{60.4} & 70.8 & \textbf{61.5} & \textbf{65.6} \\
& NeuS-Facto & \underline{40.1} & 25.7 & \underline{31.3} & \underline{56.8} & 40.8 & 47.5 & \underline{68.0} & 51.3 & 58.5 & \underline{72.2} & 58.2 & \underline{64.2} \\
& Ours &            38.6 & \textbf{27.8} & \textbf{32.3} & 55.0 & \underline{43.5} & \underline{48.6} & 66.0 & \underline{53.0 }& \underline{58.8} & 70.3 & \underline{59.2} & 64.1 \\
\midrule
\multirow{4}{*}{\rotatebox{90}{90 frames}} 
& Metashape &  \textbf{72.9} & 14.9 & 24.7 & \textbf{85.1} & 34.1 & 48.7 & \textbf{90.9} & 43.0 & 58.4 & \textbf{88.6} & 46.8 & 60.0 \\
& NeusW &        59.7 & 44.7 & 51.1 & 73.7 & \underline{59.4} & \underline{65.8} & 81.1 & 66.0 & \underline{72.8} & 81.3 & \underline{68.6} & 74.1 \\
& NeuS-Facto & \underline{64.9} & \textbf{47.0} & \textbf{54.5} & \underline{79.1} & \textbf{61.4} & \textbf{69.1} & \underline{86.4} & \textbf{69.0} & \textbf{76.7} & \underline{85.0} & 71.5 & \textbf{77.3} \\
& Ours &             60.1 & \underline{45.6} & \underline{51.8} & 73.4 & \underline{59.4} & 65.6 & 81.1 & \underline{66.1} & \underline{72.8} & 81.5 & \textbf{69.0} & \underline{74.4} \\
\bottomrule
\end{tabular}
\caption{Quantitative Comparison: 
We compare against other methods on two different settings of the Brandenburg Gate dataset. We report the precision (P), recall (R), and F1 scores. The best results are in bold, and the second best are underlined. The numbers indicate that our method is comparable to or outperforms other methods in terms of recall, especially in the 10-image setting, which is the most closed case with historical datasets. For other cases, we still achieve comparable results.
} \label{tab:comparison_quantitative}
\end{table*}
To quantitatively evaluate the 3D geometry reconstruction results, we provide precision (P), recall (R), and F1 score (F1) under three different thresholds of the generated meshes in~\cref{tab:comparison_quantitative} following the procedure of \cite{sun2022neuconw}. The precision indicates the accuracy of the reconstructed mesh compared to the ground truth mesh and the recall indicates the completeness of the results, the F1-score is a weighted score computed using precision and recall. \cref{fig:precision} illustrates these metrics. 
The three thresholds (Low, Medium, and High) correspond to 0.1, 0.2, and 0.3 meters respectively.
\begin{figure}[h]
\centering
\includegraphics[width=0.75\linewidth]{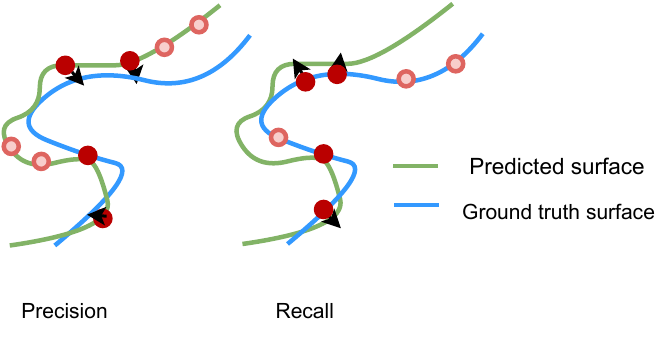}
\vspace{-0.2cm}
\caption{Demonstration of precision and recall. The red dots are under the chosen threshold, so they are used to compute the precision or recall, pink dots are over the threshold thus they are ignored.}\label{fig:precision}
\end{figure}
% Accordingly, the P, R, and F1 values are shown at three different thresholds (Low, Medium, and High) corresponding to 0.1, 0.2, and 0.3 meters respectively. 
Additionally, the area under the curve (AUC) combines all thresholds into a single metric.\par
The quantitative comparison of the Brandenburg Gate dataset shows that we achieve results comparable to other state-of-the-art methods on mesh reconstruction tasks. Our method outperforms other methods in the 10-image scenario (which is the most comparable case with historical datasets) at Low threshold for R and F1-score, which means that the mesh is more complete and closer to the ground truth mesh. Metashape~\cite{metashape2019} gains high scores for precision because it recovers the overall shape accurately. However, the mesh is incomplete bringing down the total F1 score as a result.
Note that the image quality of the Brandenburg Gate dataset is still superior to the images in historical datasets.

\begin{table*}[!h]
\centering
\begin{tabular}{ll|rrrrrrrrrrrr}
\toprule
 & & \multicolumn{3}{c}{Low} & \multicolumn{3}{c}{Medium} & \multicolumn{3}{c}{High} & \multicolumn{3}{c}{All (AUC)} \\
 \cmidrule{3-14}
& Settings & P & R & F1 & P & R & F1 & P & R & F1 & P & R & F1 \\
\midrule
\multirow{5}{*}{\rotatebox{90}{10 frames}} 
& Baseline &        37.7 & 26.6 & 31.2 & 55.4 & 44.0 & 49.1 & 67.0 & 55.0 & \underline{60.4} & 70.8 & 61.5 & 65.6 \\
& + Color loss &   37.4 & 26.7 & 31.1 & 54.9 & 43.6 & 48.6 & 66.6 & 55.0 & 60.3 & 70.0 & 61.8 & 65.4 \\
& + Sparse geo & \underline{38.9} & \underline{29.7} & \underline{33.7} & \underline{55.9} & \underline{46.6} & \underline{50.8} & \underline{67.4} & \underline{57.4} & 62.0 & \underline{71.4} & \textbf{63.2} & \underline{66.9} \\
& + Dense geo &  \textbf{42.9} & \textbf{33.0} & \textbf{37.3} & \textbf{59.5} & \textbf{49.3} & \textbf{53.9} & \textbf{69.5} & \textbf{58.3} & \textbf{63.4} & \textbf{72.9} & \underline{63.1} & \textbf{67.5} \\
& Ours &      38.6 & 27.8 & 32.3 & 55.0 & 43.5 & 48.6 & 66.0 & 53.0 & 58.8 & 70.3 & 59.2 & 64.1 \\
\midrule
\multirow{5}{*}{\rotatebox{90}{90 frames}} 
& Baseline &        59.7 & 44.7 & 51.1 & \textbf{73.7} & 59.4 & 65.8 & \underline{81.1} & 66.0 & 72.8 & 81.3 & 68.6 & 74.1 \\
& + Color loss &   \underline{60.1} & 45.3 & 51.7 & \underline{73.4} & 59.2 & 65.6 & 80.8 & 65.8 & 72.5 & 81.0 & 68.3 & 73.9 \\
& + Sparse geo & 59.5 & \underline{45.7} & 51.7 & \underline{73.4} & \underline{60.1} & \underline{66.1} & 80.9 & \underline{67.1} & \underline{73.3} & 81.1 & \underline{70.0} & \underline{74.9} \\
& + Dense geo &  \textbf{60.8} & \textbf{46.6} & \textbf{52.8} & \textbf{73.7} & \textbf{60.4} & \textbf{66.4} & \textbf{81.2} & \textbf{67.4} & \textbf{73.6} & \textbf{81.8} & \textbf{70.1} & \textbf{75.2} \\
& Ours &               \underline{60.1} & 45.6 & \underline{51.8} & \underline{73.4} & 59.4 & 65.6 & \underline{81.1} & 66.1 & 72.8 & \underline{81.5} & 69.0 & 74.4 \\
\bottomrule
\end{tabular}
\caption{Ablation Study: We report the precision (P), recall (R), and F1 scores over two different settings of the Brandenburg Gate dataset for the ablation study.  The best results are in bold, and the second best are underlined. The numbers indicate that the geometric priors, especially the dense point cloud, contribute to the reconstruction. It gives the best results in overall settings for different scores in most of the cases. Even though the color embedding appearance loss slightly deteriorates the results, it still gives comparable accuracy to the best scores.}\label{tab:brandenburg_quantitative}
\end{table*}
\input{figure_tex/ablation}
\subsection{Ablation Study} 
To analyze the influence of our proposed loss functions and the number of input images, we train our model using $10$ images and $90$ images from the Brandenburg Gate dataset where we set $90\%$ of images to gray-scale.
To quantitatively evaluate the 3D geometry reconstruction results, we also provide precision (P), recall (R), and F1 value (F1) of the generated meshes in~\cref{tab:brandenburg_quantitative}. The baseline setting is NeusW~\cite{sun2022neuconw} without geometrical loss term~\eqref{eq:geometry_loss}. We gradually add only the color appearance loss (+ color), only the geometrical loss based on two different point clouds (+ sparse geo, + dense geo), and finally our setting, \ie with color appearance loss and dense geometric loss. ~\cref{tab:brandenburg_quantitative} shows that the color embedding loss slightly degrades the geometry.

The dense point cloud supervision consistently outperforms the sparse point cloud supervision using the geometry loss.
\cref{fig:ablation} visualizes the meshes for the ablation study. The color appearance loss does not degrade the mesh qualitatively and results in small gains. That is in the 10-image case it recovers the legs of the horses and recovers one or more holes in both $10$ and $90$ images situations.
% This is equivalent to the effect of either the sparse or dense point cloud supervision.
In the case of the 10-image input (comparable to historical datasets), the sharp structure on the top left roof gets filled in the baseline case. This is almost entirely eliminated with the color appearance embedding loss.

% \newpage
\section{Conclusion}
% Reconstrucing historic datasets remains challengine. 
\paragraph{Summary} We introduced a new historical dataset that has significantly more images than previous datasets and also provide its point cloud along with camera information which is generated via SfM.
We propose a method that tackles challenges such as sparse and low-quality inputs, when reconstructing 3D shapes using archival historical datasets.
We showed that incorporating existing data such as dense point clouds can significantly improve the geometry reconstruction. The dense point cloud supervision enhances the reconstruction, especially the scenes with few images. It enables thin structures and flat, texture-less wall segments to be reconstructed and also recovers structures that are temporally changing.
Moreover, we propose a color appearance embedding loss to recover the color of the generated mesh of the historical buildings. 
% and color loss can improve the results both visually and geometrically. \david{only small details, can we say this?}
% Further feature point refinement could improve mesh
\vspace{-0.4cm}
\paragraph{Limitations and future works} The color appearance embedding loss decreases the mesh accuracy quantitatively. The capability of dealing with sparse input images still need to be improved to be able to recover detailed 3D meshes under more extreme situation. Next, we plan to explore methods that are especially targeted at few-shot view synthesis~\cite{ajay2021, yang2023freenerf} and reconstruction methods.
\newpage

%%%%%%%%% REFERENCES
{\small
\nocite{*}
\bibliographystyle{ieee_fullname}
\bibliography{egbib}
}

\onecolumn
\appendix
% \section{Appendix}

\renewcommand \thesection{A\arabic{section}}
\renewcommand{\thefigure}{A.\arabic{figure}}
\renewcommand{\thetable}{A.\arabic{table}}
% This Supplement contains information on code, datasets, some more results. 

\section{Used Code and Datasets}
\cref{tab:code_data} summarizes the code and datasets we use for evaluation and comparison. Our code and recorded datasets will be made publicly available upon acceptance.
\begin{table}[ht!]
    \centering
    \begin{tabular}{@{}lp{3cm}llp{7cm}c@{}}
    \toprule
         & Name & type & year & link & license \\
    \midrule
    \cite{metashape2019} & Metashape & code & 2021 & \small{\url{https://www.agisoft.com/}} & Proprietary \\
    \cite{sun2022neuconw} & NeuralRecon-W & code & 2022 & \small{\url{https://github.com/zju3dv/NeuralRecon-W}} & Apache-2.0 \\
    % \cite{nerfstudio} & NeRFacto? & code & 2022 & \small{\url{https://docs.nerf.studio/nerfology/methods/nerfacto.html}} & Apache-2.0 \\
    \cite{Yu2022SDFStudio} & NeuS-Facto, NeusW & code & 2022 & \small{\url{https://docs.nerf.studio/nerfology/methods/nerfacto.html}} & Apache-2.0 \\
    \cite{colmap} & COLMAP & code & 2016 & \small{\url{https://colmap.github.io/}} & new BSD \\
    \cite{detone18superpoint,lindenberger2023lightglue,tyszkiewicz2020disk} & Hierarchical Localization Toolbox
 & code & 2023 & \small{\url{https://github.com/cvg/Hierarchical-Localization/}} &  Apache-2.0 \\
    \cite{brandenburg} & Brandenburg Gate & dataset & 2020 & \small{\url{https://www.cs.ubc.ca/~kmyi/imw2020/data.html}} & - \\
    \cite{jena} & Historic Building & dataset & 2022  & \small{\url{https://www.gw.uni-jena.de/en/faculty/juniorprofessur-fuer-digital-humanities/research/jena4d-stadtgeschichtsbuch}} & Creative Commons \\
    Ours & National Theater & dataset & 2023 & will be public upon acceptance & Creative Commons \\
    \bottomrule
    \end{tabular}
    \vspace{0.25cm}
    \caption{Used datasets and code in our submission, together with reference, link, and license.}
    \label{tab:code_data}
\end{table}
For the pre-processing of the dataset, \ie, mask out humans and irrelevant objects, we use the NeuralRecon-W~\cite{sun2022neuconw} codebase. For NeusW we used the SDF-Studio implementation. 
\section{Mesh Visualization}
We show in this section the reconstructed meshes of Brandenburg Gate~\cite{brandenburg} dataset, corresponding to~\cref{tab:comparison_quantitative}. 
\begin{figure}[th!]
\centering
\setlength{\tabcolsep}{3pt}
\newcolumntype{Y}{>{\centering\arraybackslash}p{0.151\linewidth}}
\begin{tabular}{lYYYYYY}
&Ground truth & Metashape~\cite{metashape2019} & Neus-Facto~\cite{Yu2022SDFStudio} & NeusW~\cite{sun2022neuconw} & Ours & Ours color \\
\multirow{2}{*}{\rotatebox{90}{10 frames}} &
\includegraphics[width=\linewidth, clip, viewport=270 0 699 348,angle=180,origin=c]{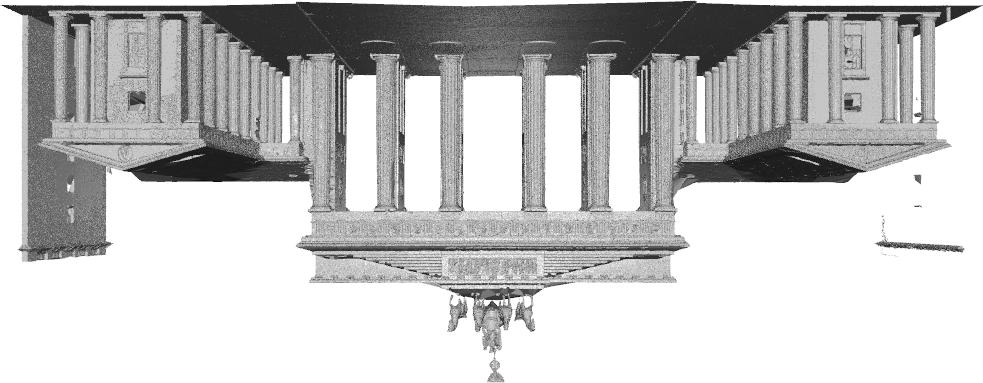} &
\includegraphics[width=\linewidth,angle=180,origin=c]{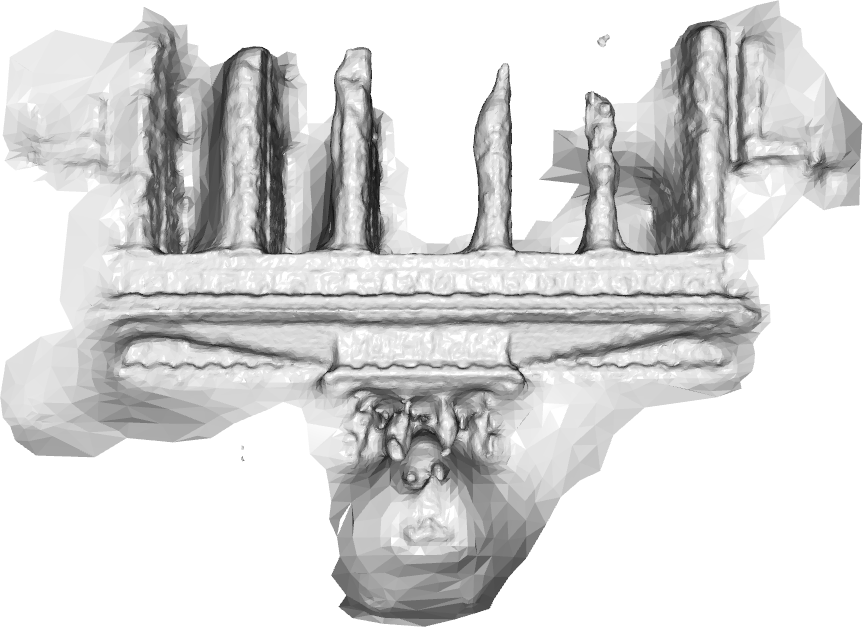} &
\includegraphics[width=\linewidth, clip, viewport=270 50 699 398]{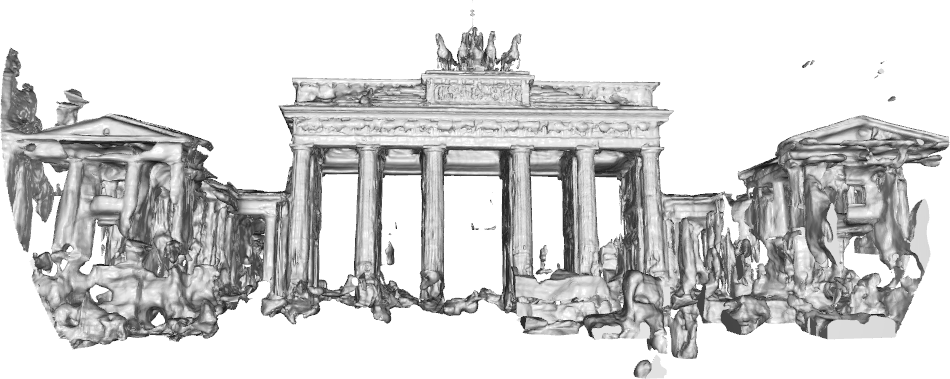} &
\includegraphics[width=\linewidth, clip, viewport=250 0 679 348,angle=180,origin=c]{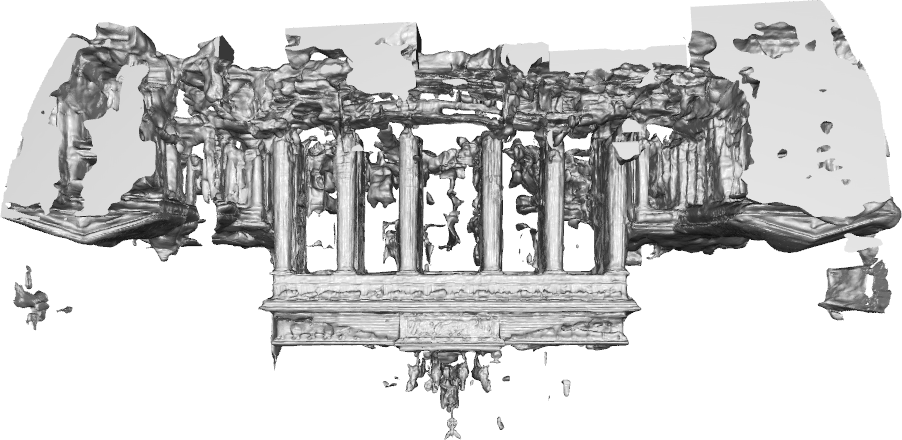} & 
\includegraphics[width=\linewidth, clip, viewport=270 0 699 348,angle=180,origin=c]{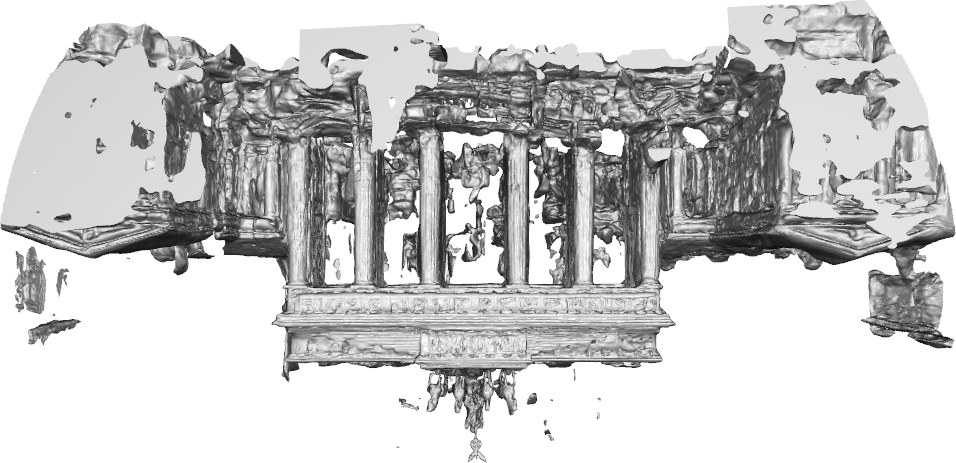} & 
\includegraphics[width=\linewidth, clip, viewport=270 0 699 348,angle=180,origin=c]{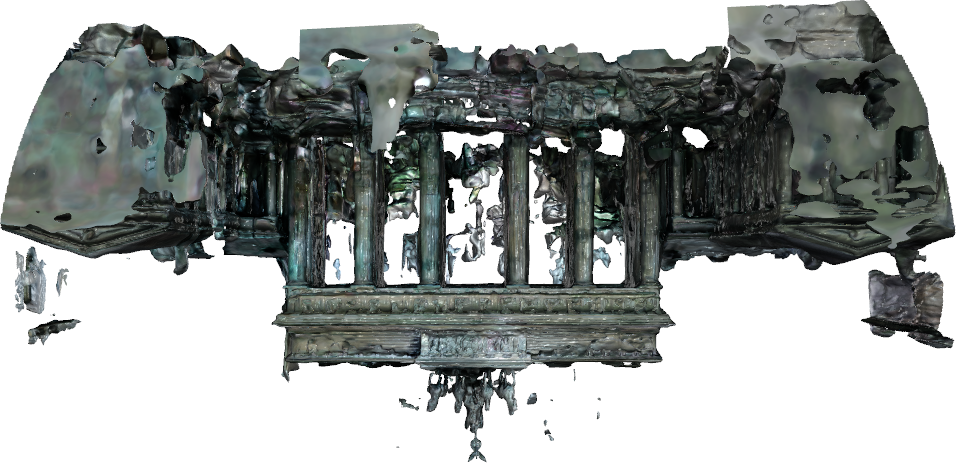} \\
&
\includegraphics[width=\linewidth, clip, viewport=270 0 699 348,angle=180,origin=c]{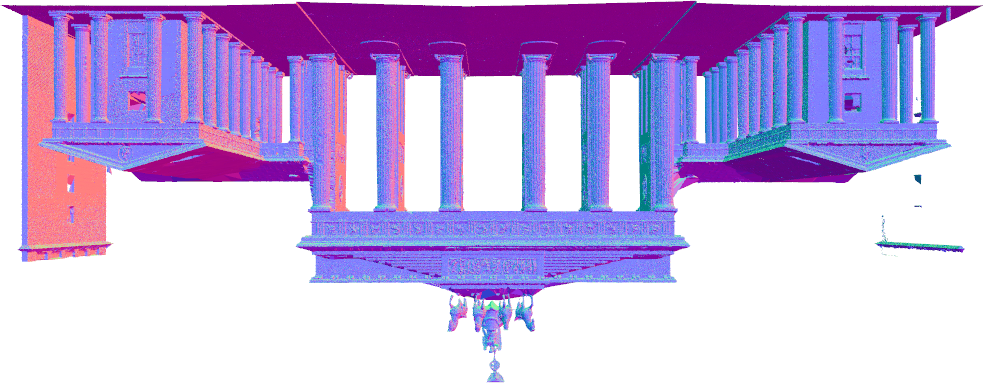} &
\includegraphics[width=\linewidth,angle=180,origin=c]{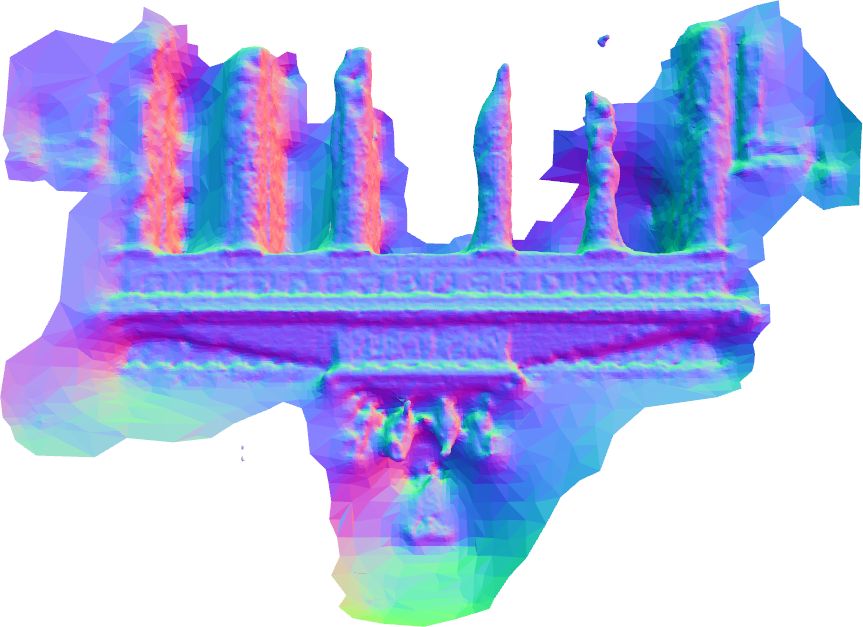} &
\includegraphics[width=\linewidth, clip, viewport=270 50 699 398]{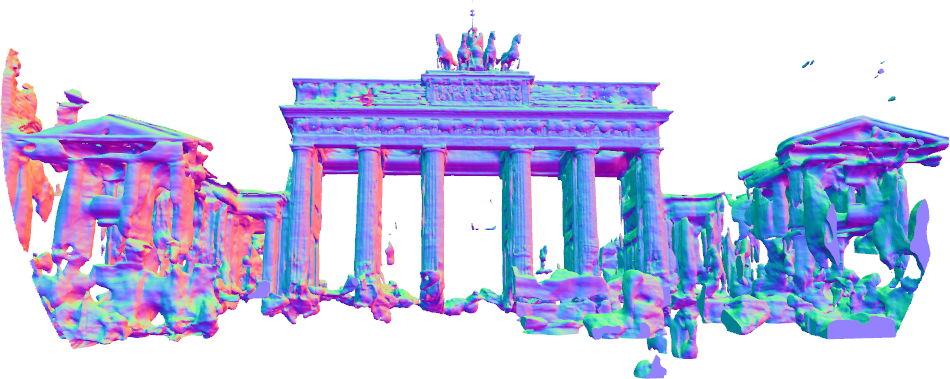} &
\includegraphics[width=\linewidth, clip, viewport=250 0 679 348,angle=180,origin=c]{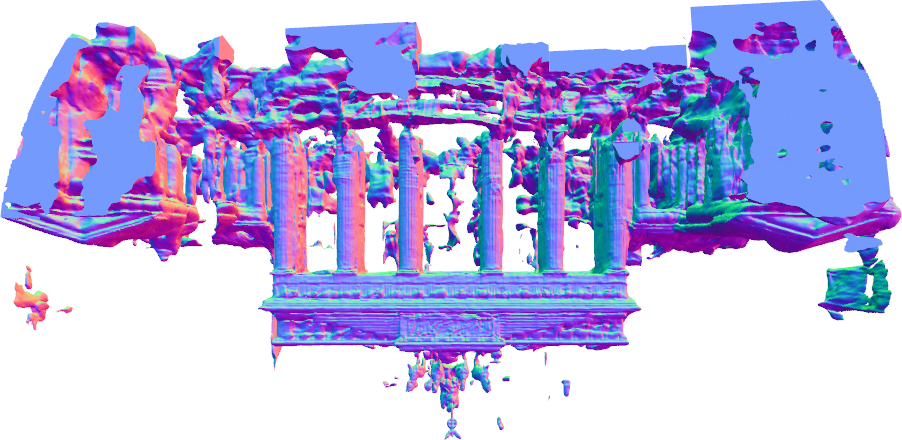} & 
\includegraphics[width=\linewidth, clip, viewport=270 0 699 348,angle=180,origin=c]{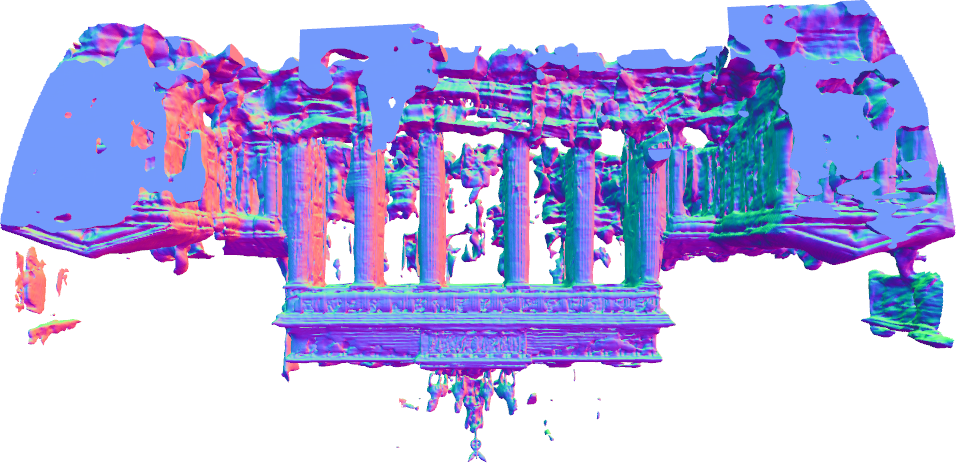} & 
\includegraphics[width=\linewidth, clip, viewport=400 0 559 150,angle=180,origin=c]{imgs/ablation/supp/10_ours2_normal.png} \\
\multirow{2}{*}{\rotatebox{90}{90 frames}} &
\includegraphics[width=\linewidth, clip, viewport=270 0 699 348,angle=180,origin=c]{imgs/ablation/gate_90/gt_no_color_crop.png} &
\includegraphics[width=\linewidth,clip, viewport=300 0 729 348,  angle=180,origin=c]{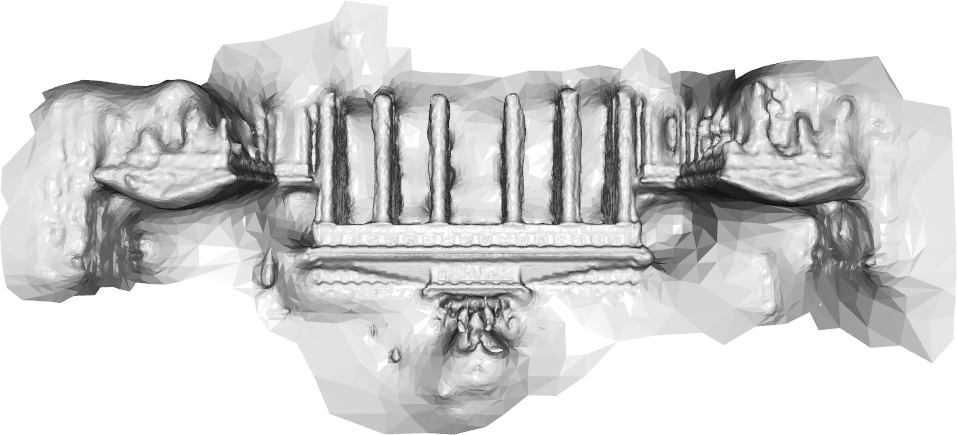} &
\includegraphics[width=\linewidth, clip, viewport=270 50 699 398]{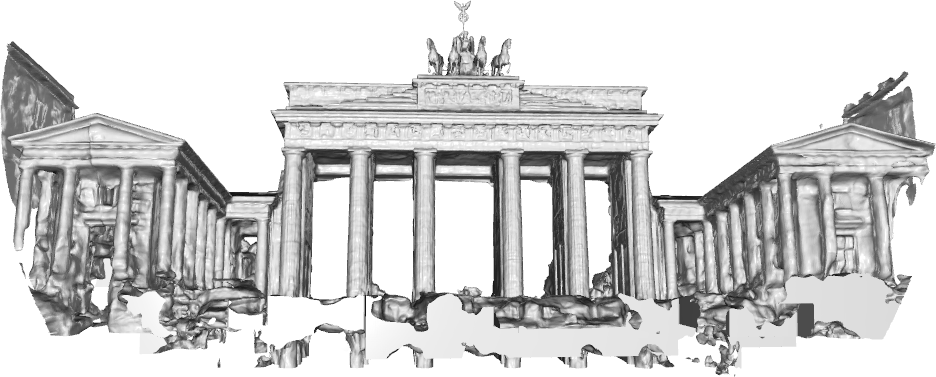} &
\includegraphics[width=\linewidth, clip, viewport=250 0 679 348,angle=180,origin=c]{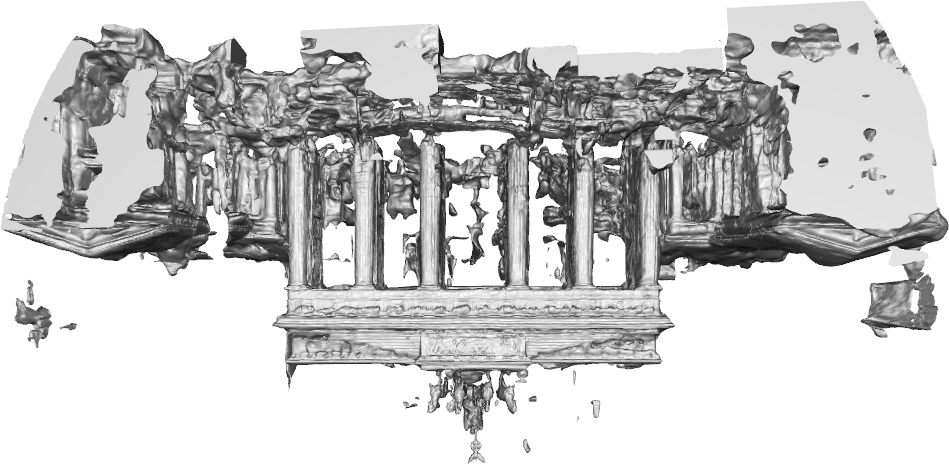} & 
\includegraphics[width=\linewidth, clip, viewport=270 0 699 348,angle=180,origin=c]{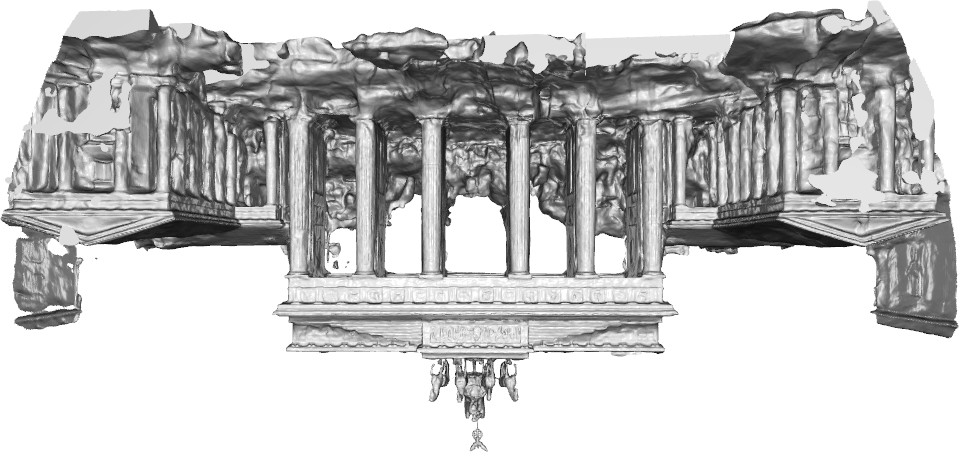} & 
\includegraphics[width=\linewidth, clip, viewport=270 0 699 348,angle=180,origin=c]{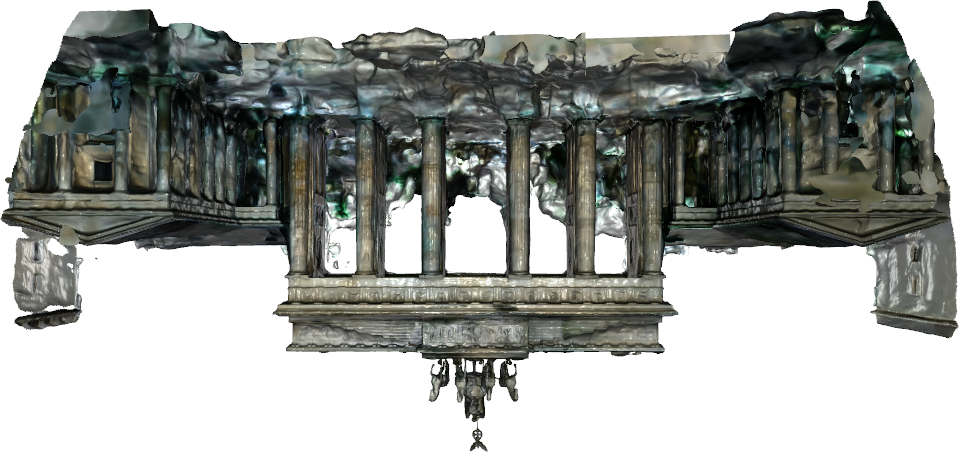} \\
& \includegraphics[width=\linewidth, clip, viewport=270 0 699 348,angle=180,origin=c]{imgs/ablation/gate_90/gt_normal_crop.png} &
\includegraphics[width=\linewidth,clip, viewport=300 0 729 348,  angle=180,origin=c]{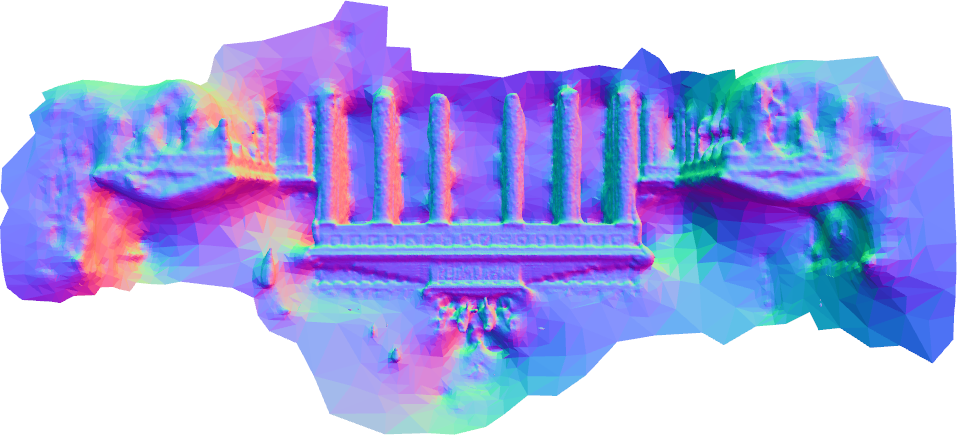} &
\includegraphics[width=\linewidth, clip, viewport=270 50 699 398]{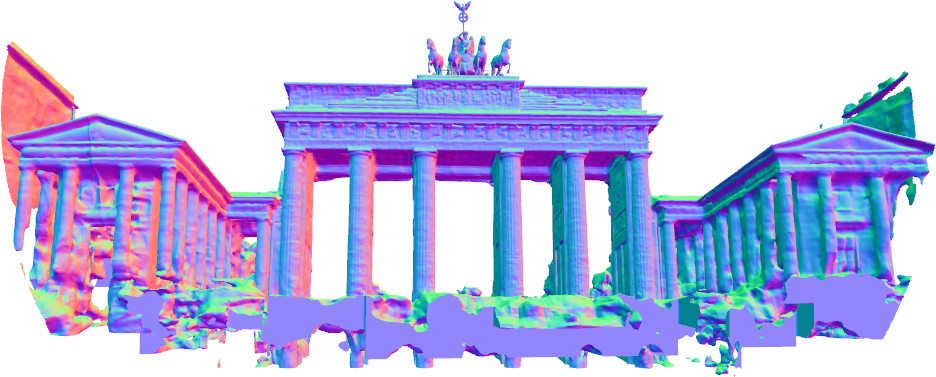} &
\includegraphics[width=\linewidth, clip, viewport=250 0 679 348,angle=180,origin=c]{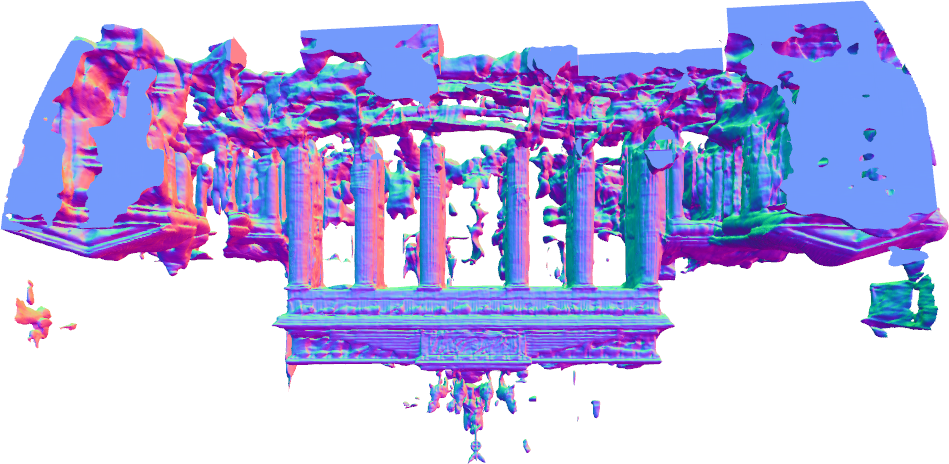} & 
\includegraphics[width=\linewidth, clip, viewport=270 0 699 348,angle=180,origin=c]{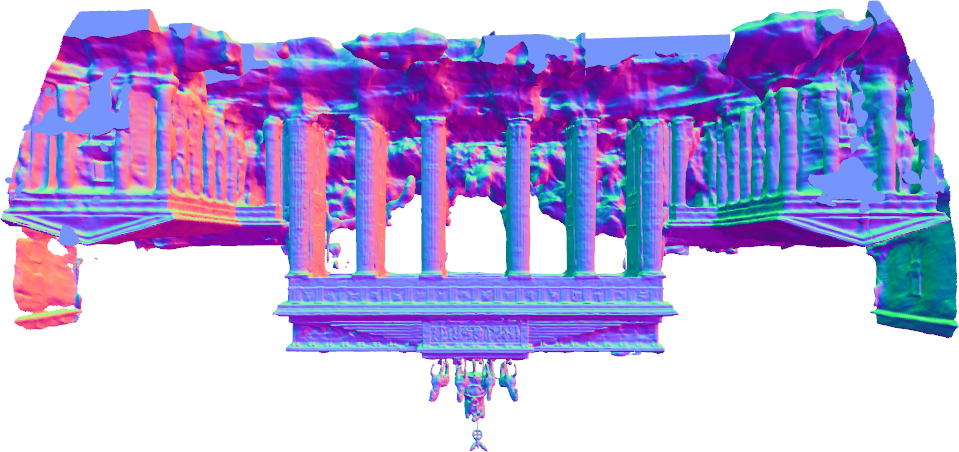} & 
\includegraphics[width=\linewidth, clip, viewport=400 0 559 150,angle=180,origin=c]{imgs/ablation/supp/90_ours_normal.png}
\end{tabular}
\caption{Reconstructed meshes comparison results for Brandenburg Gate~\cite{brandenburg} dataset.}\label{fig:supp_brandenburg}
\vspace{-0.3cm}
\end{figure}

\section{Rendering Visualization}

\begin{figure}[ht!]
\centering
\setlength{\tabcolsep}{3pt}
\newcolumntype{Y}{>{\centering\arraybackslash}p{0.28\textwidth}}
\begin{tabular}{cYYYY}
    & Ground Truth & Baseline & Ours \\
    \rotatebox{90}{10 frames} & 
    \includegraphics[width=\linewidth]{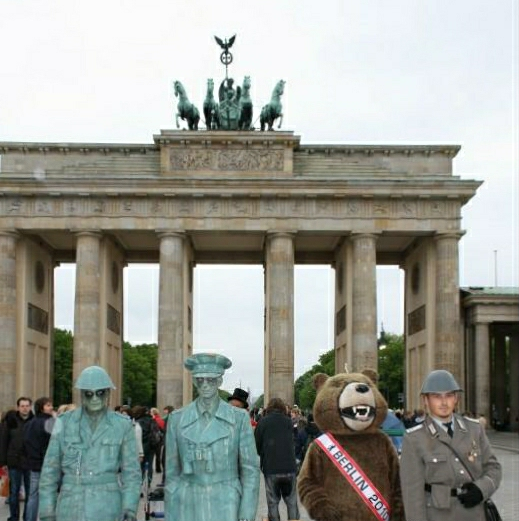} &
    \includegraphics[width=\linewidth]{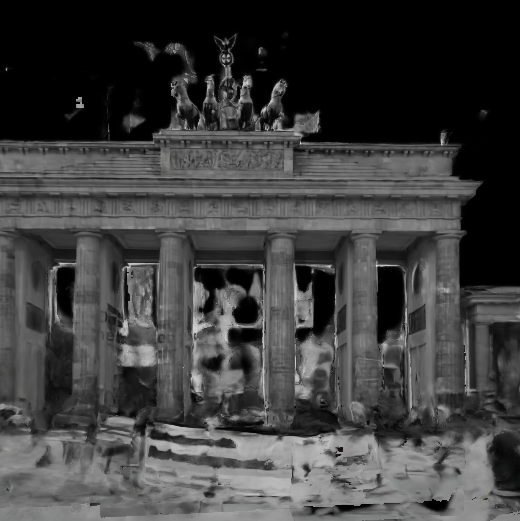} &
    \includegraphics[width=\linewidth]{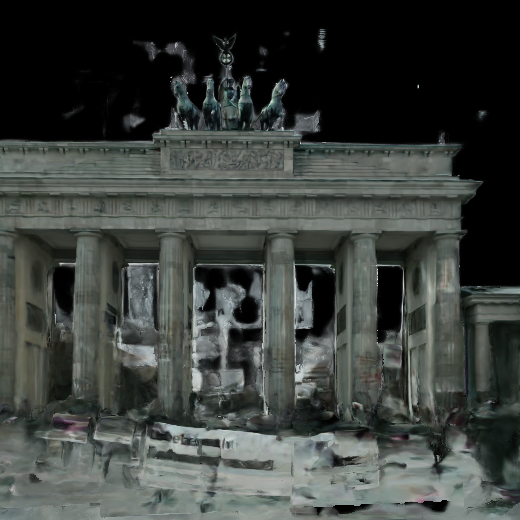} 
    \\
    \rotatebox{90}{90 frames} & 
    \includegraphics[width=\linewidth]{imgs/supplementary/brandenburg/neusW-brandenburg_gate_historic-gt.png} &
    \includegraphics[width=\linewidth]{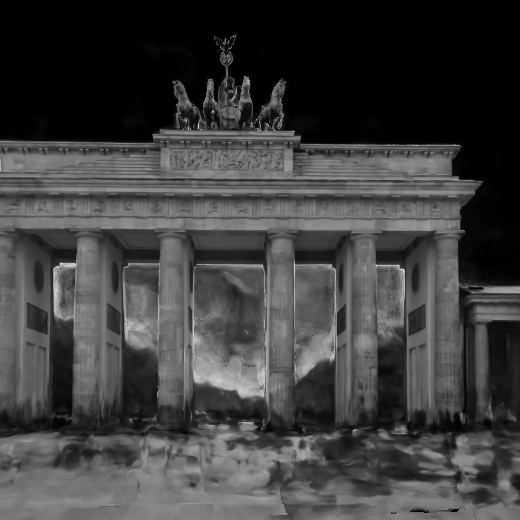} &
    \includegraphics[width=\linewidth]{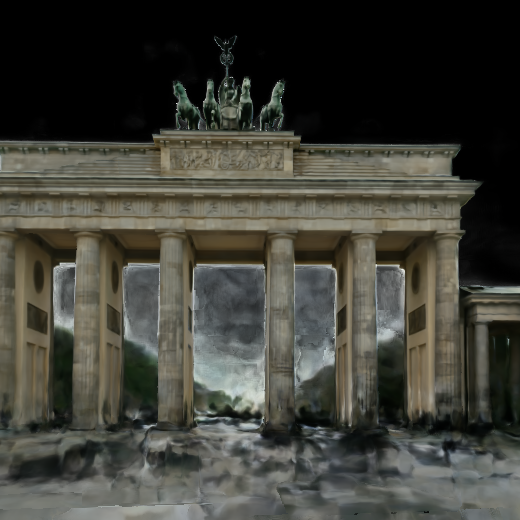} \\
  \end{tabular}
  \caption{Comparison results of novel view synthesis of Brandenburg Gate for baseline NeusW~\cite{sun2022neuconw} and our method. With the color appearance embedding loss, we can recover the color images as well.}
  \label{fig:color_rendering_brandenburg}
  \vspace{-0.3cm}
\end{figure}

In this section, we show the rendered view, which is the by-product of our method. We use volumetric rendering to get high-fidelity surface which does not need root-finding when extracting meshes.   %no artificial lighting is added, 
% we can gen, and the resolution of marching cubes does not affect fine details. The downside is that real-time rendering is not possible.
In~\cref{fig:color_rendering_brandenburg} we show the results with and without color loss on the Brandenburg Gate dataset corresponding to~\cref{fig:ablation} (the second and the last columns) and~\cref{tab:brandenburg_quantitative}.
The network can already recover the colors with only one color image.

% (Average appearance embedding, multiple time of day photos influence appearance but color is there)

% The rendering also contains some cloud artifacts between the columns. These are accumulated during rendering and not part of the extracted mesh.

\begin{figure}[hb!]
\centering
\setlength{\tabcolsep}{3pt}
\newcolumntype{Y}{>{\centering\arraybackslash}p{0.22\textwidth}}
\begin{tabular}{lYYYY} 
    & GT & Color & Normal  & Depth \\
    \rotatebox{90}{Theater} & 
    \includegraphics[width=\linewidth]{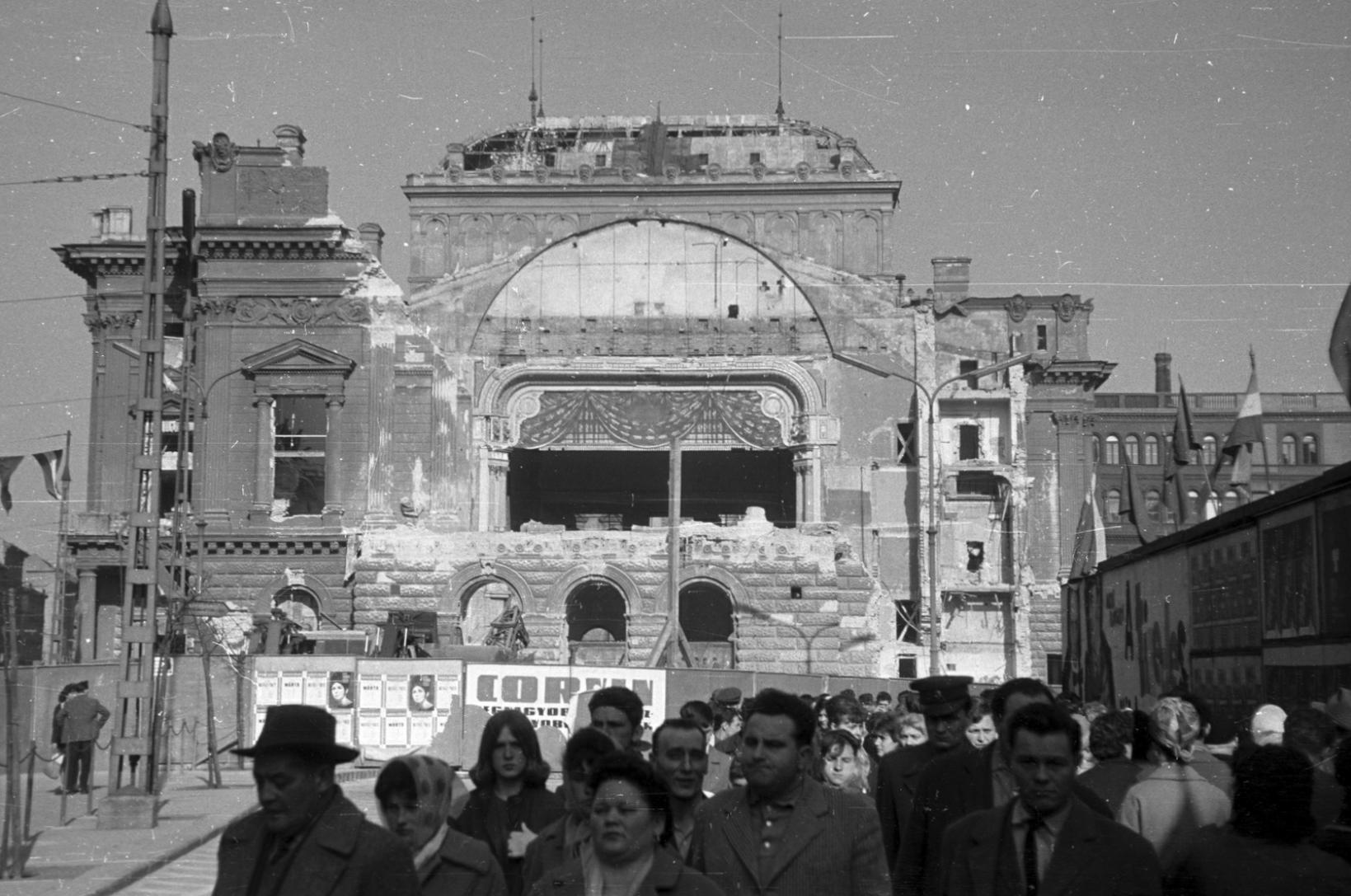} &
    \includegraphics[width=\linewidth]{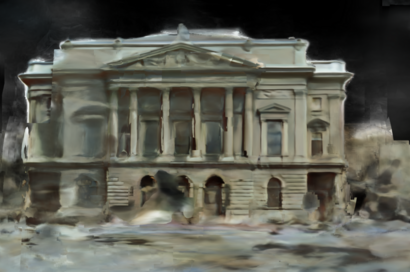} &
    \includegraphics[width=\linewidth]{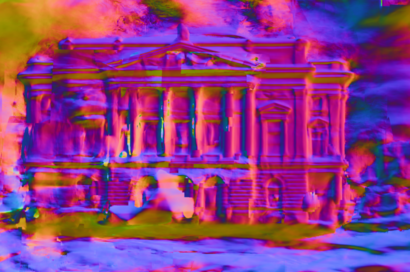} &
    \includegraphics[width=\linewidth]{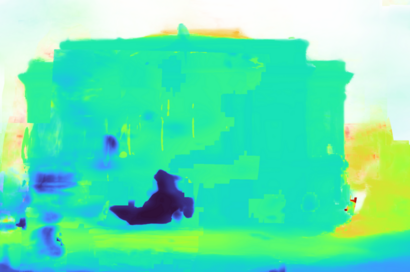} 
    \\
    \rotatebox{90}{Observatory} & 
    \includegraphics[width=\linewidth]{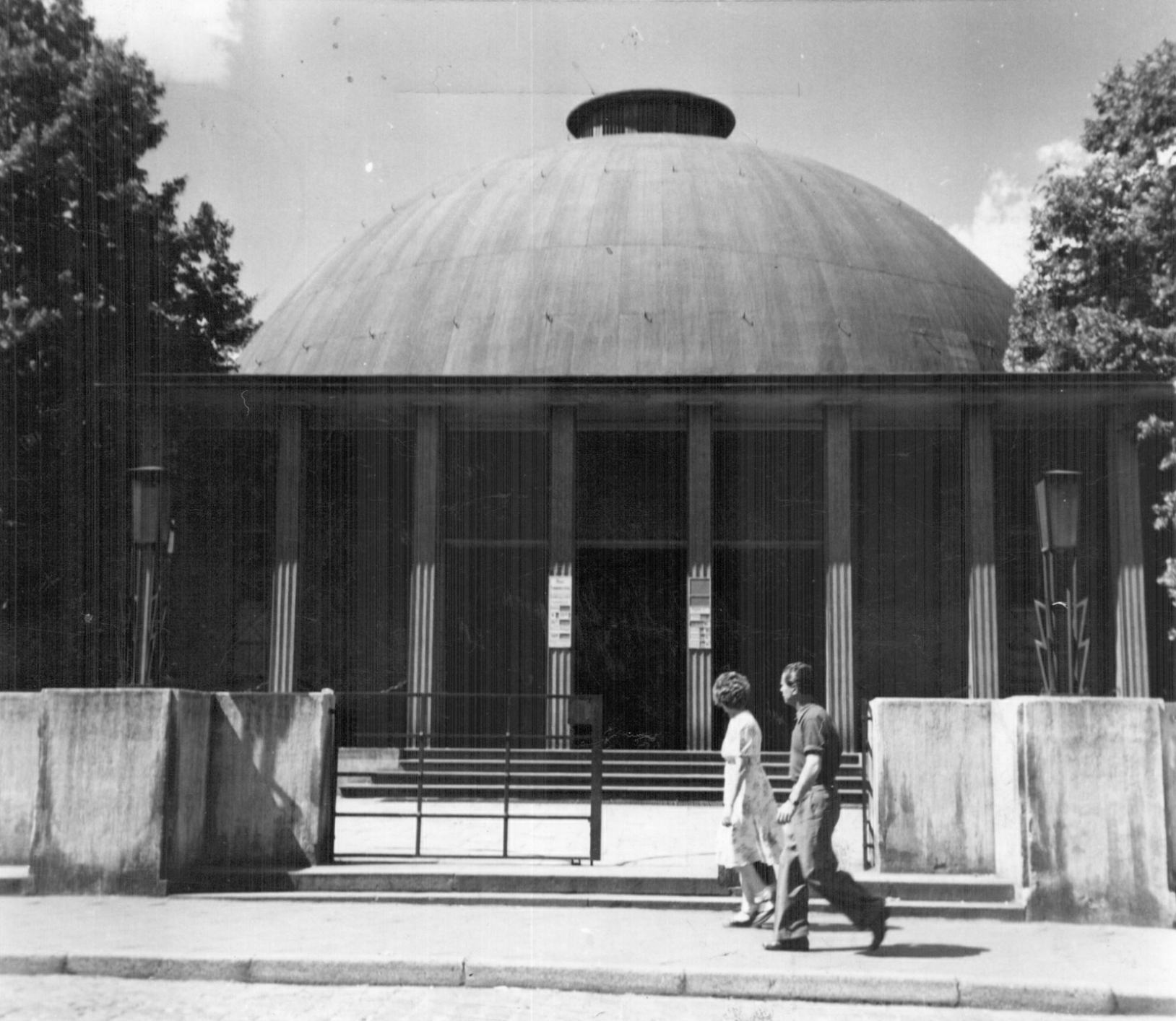} &
    \includegraphics[width=\linewidth]{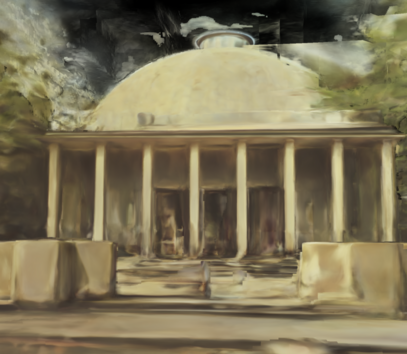} &
    \includegraphics[width=\linewidth]{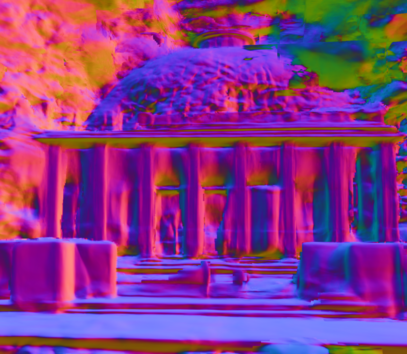} &
    \includegraphics[width=\linewidth]{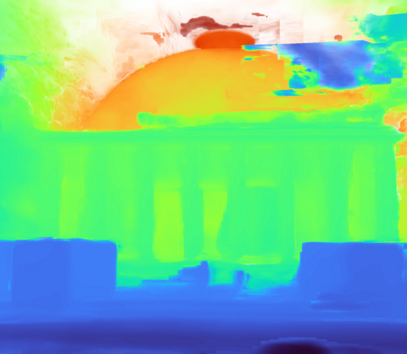} 
    \\
    \rotatebox{90}{\small Hotel International} & 
    \includegraphics[width=\linewidth]{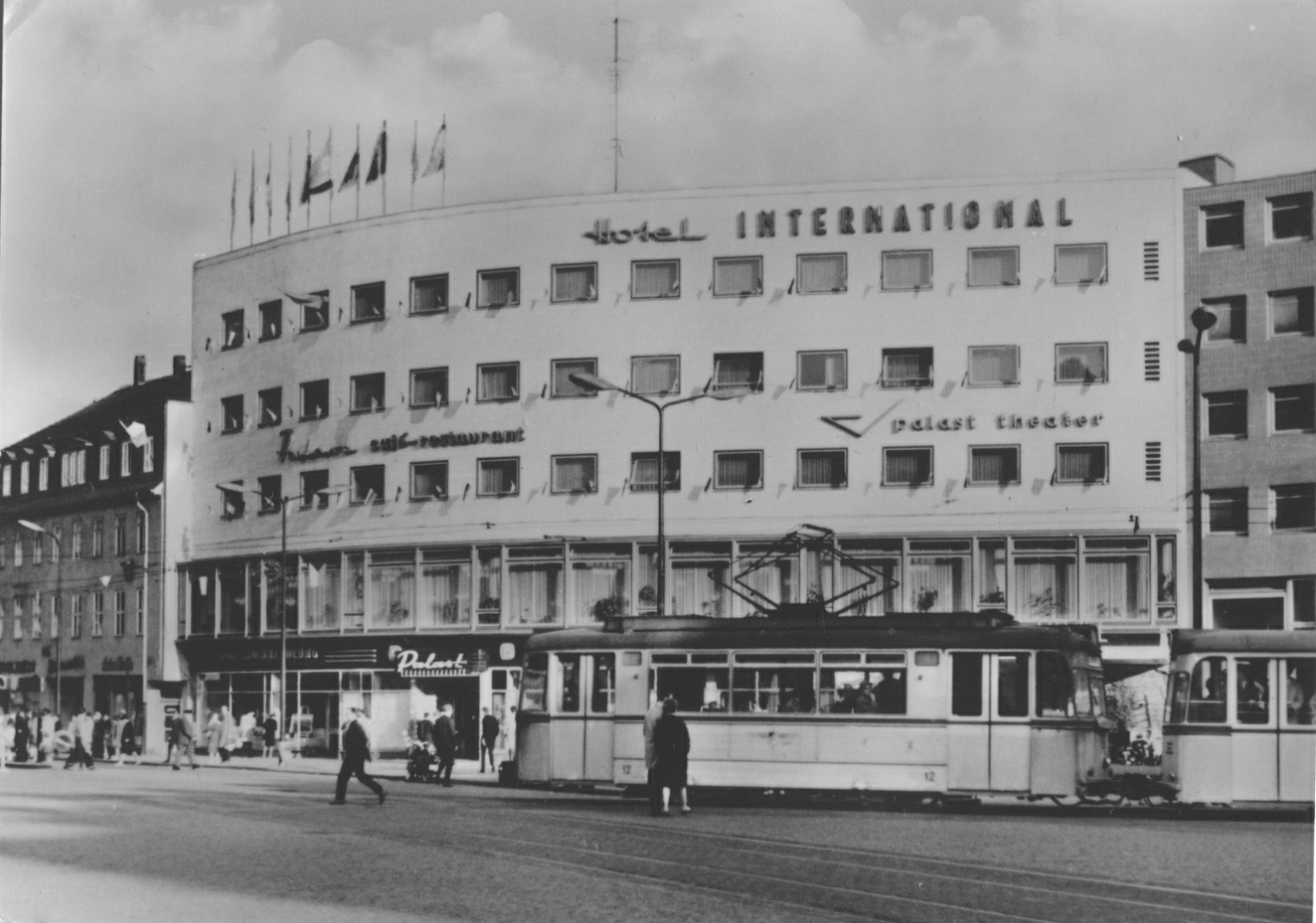} &
    \includegraphics[width=\linewidth]{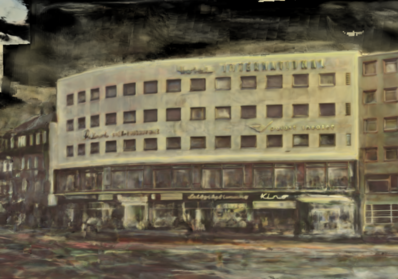} &
    \includegraphics[width=\linewidth]{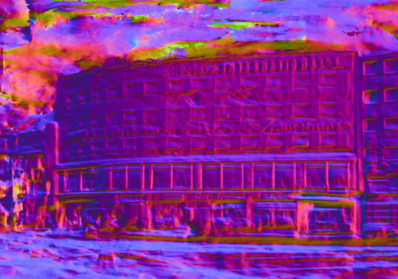} &
    \includegraphics[width=\linewidth]{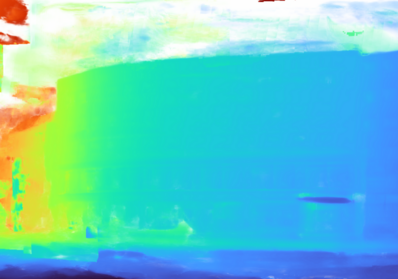} 
    \\
    \rotatebox{90}{\small St\. Michael Church} & 
    \includegraphics[width=\linewidth]{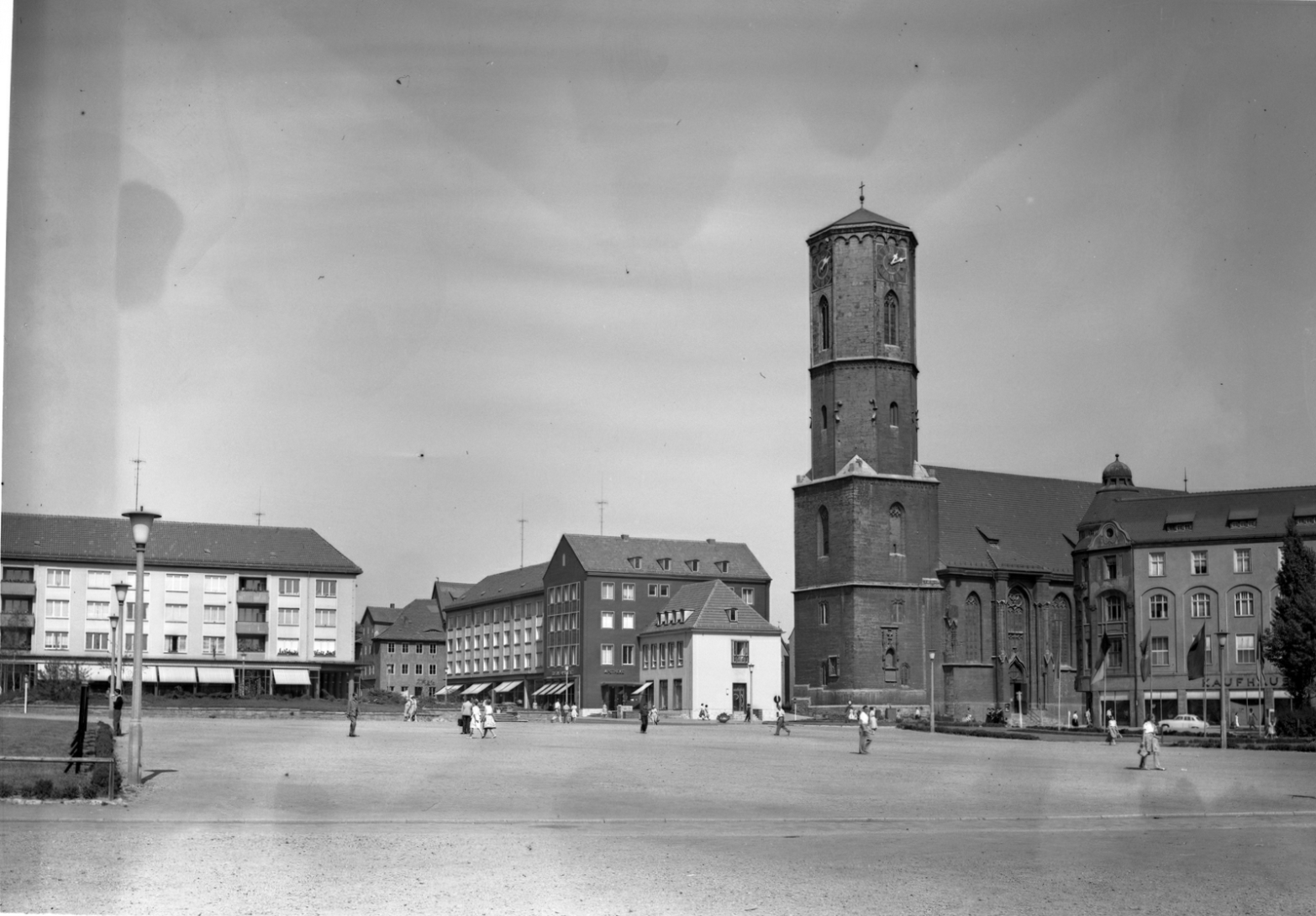} &
    \includegraphics[width=\linewidth]{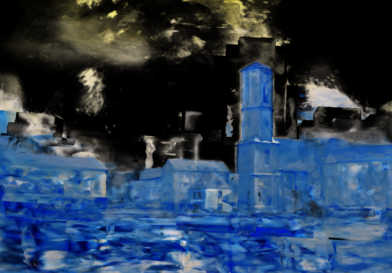} &
    \includegraphics[width=\linewidth]{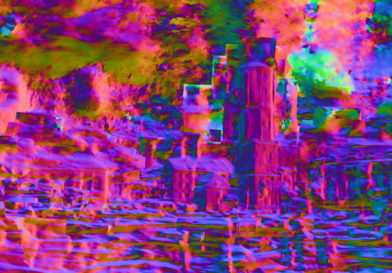} &
    \includegraphics[width=\linewidth]{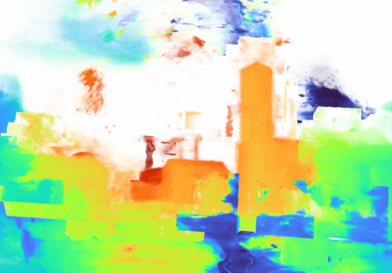} 
  \end{tabular}
  \caption{Novel View Synthesis of historical datasets. The first column shows the ground truth view, the second column the rendered color images, the third and fourth columns are normal and depth maps. The normal maps reflect more details for each ray than normal from meshes. The normal of the meshes are restricted by the Marching Cubes resolutions.}
  \label{fig:color_rendering_historic}
\end{figure}

\cref{fig:color_rendering_historic} shows the rendered images, depth, and normal maps of our method on the historical datasets corresponding to the last column of~\cref{fig:mesh_results}. Note that the depth and normal maps are rendered directly without extracting meshes, different from the mesh normal we show in~\cref{fig:mesh_results}. We are able to recover the color images as well, despite the limited color input images. For the St. Micheal Church dataset, however, the recovered color is less successful compared to the other historical datasets due to the fact that the dataset does not contain any color images. The color is only projected to gray-scale using the perceptual weights without color supervision. 
\end{document}